%% file: template.tex
\documentclass{article}

\pdfoutput=1

\usepackage{arxiv}

\usepackage{times}
\usepackage{latexsym}

\usepackage[T1]{fontenc}
\usepackage[utf8]{inputenc}

\usepackage{microtype}

\usepackage{inconsolata}

\usepackage{listings}
\usepackage{wrapfig}
\usepackage{hyperref}       
\usepackage{url}            
\usepackage{booktabs}      
\usepackage{amsfonts}      
\usepackage{nicefrac}       
\usepackage{microtype}      
\usepackage{xcolor}        
\usepackage{amsmath}
\usepackage{amssymb}
\usepackage[section]{placeins}
\usepackage{graphicx}
\usepackage{natbib}
\usepackage{booktabs}
\usepackage{tabularx}

\usepackage{adjustbox}
\usepackage{xspace}
\usepackage{colortbl} %\cellcolor
\usepackage{soul} %\sethlcolor
\usepackage{tabularx}
\usepackage{stfloats}
\usepackage{algorithm,algorithmicx}
\usepackage[noend]{algpseudocode}
\usepackage{algcompatible}
\usepackage{pifont}% http://ctan.org/pkg/pifont
\usepackage{makecell}
\graphicspath{{./figures/}}
\usepackage{multirow}
\usepackage{enumitem}

\definecolor{lightgrey}{HTML}{dcdbdb}
\sethlcolor{lightgrey}
\definecolor{lightblue}{HTML}{E8F0FE}
\definecolor{lightblue}{HTML}{E8F0FE}
\definecolor{gray}{HTML}{9aa0a6}
\definecolor{lightpink}{HTML}{F48FB1}
\definecolor{lightred}{HTML}{FFCBC9}
\definecolor{lightcyan}{HTML}{80DEEA}
\definecolor{darkgreen}{rgb}{0.0, 0.5, 0.0}
\definecolor{babyblueeyes}{rgb}{0.1, 0.3, 0.5}
\definecolor{turquoisegreen}{rgb}{0.3, 0.6, 0.3}

\usepackage[skins]{tcolorbox} 
\tcbuselibrary{breakable} 
\newtcolorbox[auto counter, number within=section, list type=subsubsection, list inside=toc]{sectionbox}[2][]{
colback=white!98!gray, colframe=black, 
colbacktitle=white!90!gray, coltitle=black, 
fonttitle=\bfseries,
title={#2}, 
list entry={Comment \thetcbcounter\quad}
}
\usepackage{lipsum}
\usepackage{soul}

\definecolor{greengrey}{rgb}{0.0, 0.3, 0.0} % This is a base darkgreen color
\colorlet{greengreywithhint}{greengrey!90!gray} % Mixing 80% darkgreen with 20% gray

\newcommand{\method}{\textbf{ExeSQL}}
\title{\method: Self-Taught Text-to-SQL Models with Execution-Driven Bootstrapping for SQL Dialects}

\usepackage{multirow}
\usepackage{colortbl}
\usepackage{graphicx}
\usepackage{makecell} % for \Xhline to customize thickness of rules

\author{Jipeng Zhang$^1$\footnotemark[1]\protect\phantom{\footnotesize 1}, Haolin Yang$^1$\footnotemark[1]\protect\phantom{\footnotesize 1}, Kehao Miao$^{2}$\thanks{\, Equal Contribution. Code are available at the following links: \url{https://github.com/2003pro/exesql}.}\protect\phantom{\footnotesize 1}, \textbf{Ruiyuan Zhang}$^1$, Renjie Pi$^1$, Jiahui Gao$^3$, Xiaofang Zhou$^1$ \\
\\
  $^1$The Hong Kong University of Science and Technology\\
  $^2$Nanyang Technological University, $^3$The University of Hong Kong \\
\texttt{\{jzhanggr,hyangby,zry,rpi,zxf\}}@ust.hk\\ 
\texttt{kehao001@e.ntu.edu.sg}, \texttt{ggaojiahui@gmail.com}
}

\renewcommand{\shorttitle}{}

\hypersetup{
pdftitle={A template for the arxiv style},
pdfsubject={q-bio.NC, q-bio.QM},
pdfauthor={David S.~Hippocampus, Elias D.~Striatum},
pdfkeywords={First keyword, Second keyword, More},
}

\begin{document}
\maketitle
\begin{abstract}

% Recent text-to-SQL models have achieved impressive performance, but progress has been largely confined to SQLite due to dataset limitations. However, real-world enterprise text-to-SQL applications often involve many SQL dialects with diverse syntax and functions, a challenge overlooked by current LLMs. The primary challenge in developing a dialect-aware LLM lies in acquiring high-quality dialect-specific text-to-SQL datasets and effectively integrating dynamic execution-based signals beyond static ``text and SQL'' strings. To bridge this gap, we propose an execution-driven solution comprising dialect dataset generation, execution-guided bootstrapping, preference data collection, and a two-stage fine-tuning framework that combines supervised finetuning with preference learning. Our results show that this method significantly outperforms existing models, achieving improvements of 15.2\% and 13.21\% over the GPT-4 baseline in 2 different SQL dialects.

Recent text-to-SQL models have achieved strong performance, but their effectiveness remains largely confined to SQLite due to dataset limitations. However, real-world applications require SQL generation across multiple dialects with varying syntax and specialized features, which remains a challenge for current models. 
%The main obstacle in building a dialect-aware model lies in acquiring high-quality dialect-specific data and integrating execution feedback beyond static text-SQL pairs. This work introduces ExeSQL, a text-to-SQL framework with execution-driven bootstrapping. The approach consists of agentic translation bootstrapping, iterative data generation, and iterative preference training, allowing models to adapt to different SQL dialects through execution-guided learning. Experiments show that ExeSQL bridges the dialect gap in text-to-SQL, achieving average improvements of 15.2\%, 10.38\%, and 4.49\% over GPT-4o across three distinct SQL dialects (PostgreSQL, MySQL, Oracle).
The main obstacle in building a dialect-aware model lies in acquiring high-quality dialect-specific data. Data generated purely through static prompting—without validating SQLs via execution—tends to be noisy and unreliable. Moreover, the lack of real execution environments in the training loop prevents models from grounding their predictions in executable semantics, limiting generalization despite surface-level improvements from data filtering. This work introduces ExeSQL, a text-to-SQL framework with execution-driven, agentic bootstrapping. The method consists of iterative query generation, execution-based filtering (e.g., rejection sampling), and preference-based training, enabling the model to adapt to new SQL dialects through verifiable, feedback-guided learning.
Experiments show that ExeSQL bridges the dialect gap in text-to-SQL, achieving average improvements of 15.2\%, 10.38\%, and 4.49\% over GPT-4o on PostgreSQL, MySQL, and Oracle, respectively, across multiple datasets of varying difficulty.

\end{abstract}

\section{Introduction}
% \begin{figure}[h]
%     \centering
%     \includegraphics[width=0.48\textwidth]{figures/new_motivation.pdf}
%     \caption{Given a natural language question, different SQL dialects require distinct syntax adjustments, such as explicit type casting in PostgreSQL. Beyond the traditional text-input–SQL-output formulation, we incorporate the database environment to enable agentic execution feedback for data synthesis and training.}
%     \label{fig:motivation}
% \end{figure}

\begin{wrapfigure}{r}{0.48\textwidth}
  \vspace{-4mm}
  \centering
  \includegraphics[width=0.95\linewidth]{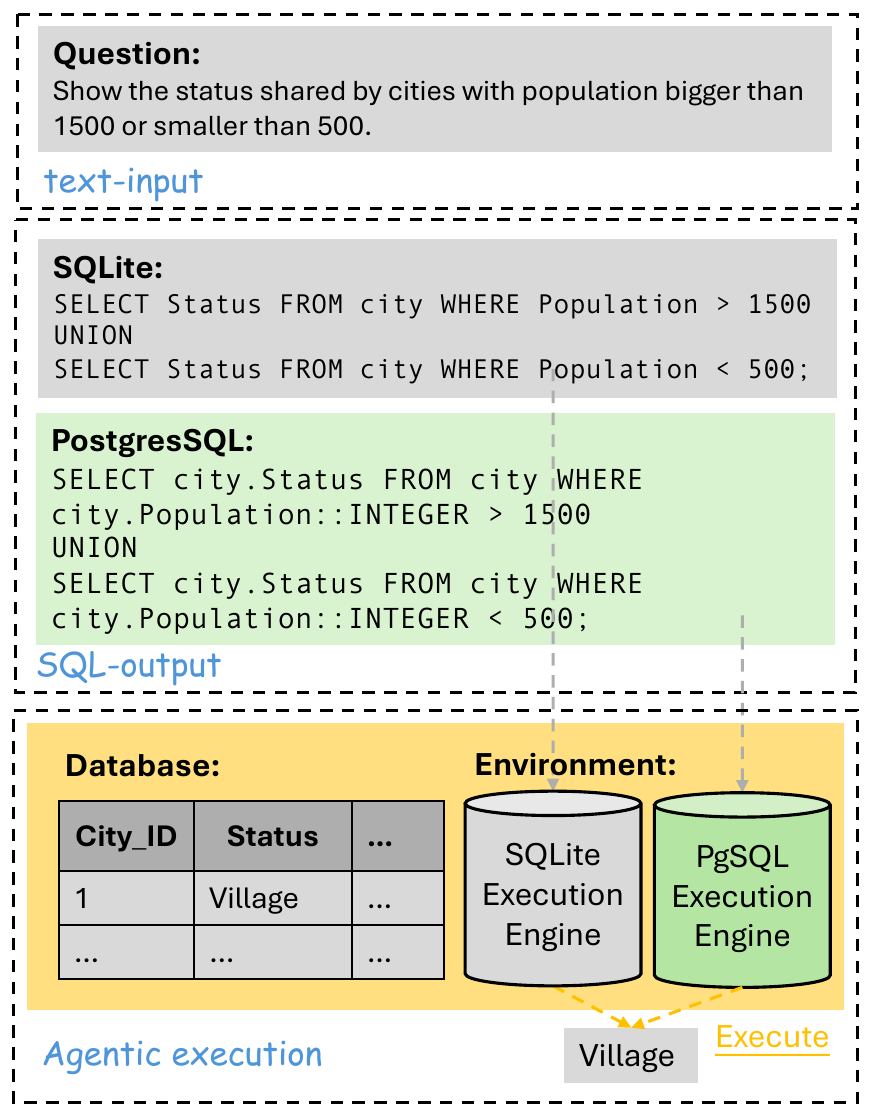}
  \vspace{-3mm}
  \caption{
    Given a natural language question, different SQL dialects require distinct syntax adjustments, such as explicit type casting in PostgreSQL. Beyond the traditional text-input–SQL-output formulation, we incorporate the database environment to enable agentic execution feedback for data synthesis and training.
  }
  \label{fig:motivation}
  \vspace{-6mm}
\end{wrapfigure}

% 记得重写这部分强调一下text-to-sql和semantic parsing的关系
With the rapid advancement of Large Language Models (LLMs), their capabilities in assisting with various coding tasks have significantly improved. Tools like GitHub Copilot~\citep{copilot,whisper-copilot} and models such as OpenAI Codex~\citep{chen2021humaneval} have enhanced developer productivity by automating repetitive tasks, providing real-time suggestions, and offering detailed explanations of code functionality. One crucial application of LLMs in software development is the automatic generation of SQL queries from text (text-to-SQL), a task that has gained increasing attention~\citep{wikisql2017,yu2018spider,li2024bird,spider2}. However, most existing research~\citep{li2024codes,zhuang2024structlm,dong2023c3sql,2023-din-sql,wang2023macsql,gan-etal-2021-natural-sql,deng2021spiderrealistic} and datasets in the text-to-SQL domain are primarily designed for SQLite, with limited coverage of widely used database systems such as MySQL, PostgreSQL, BigQuery, Oracle, and DuckDB. We incorporate an example of a question with dialect SQL in Figure~\ref{fig:motivation}. The lack of high-quality, dialect-specific text-to-SQL data presents significant challenges in developing models that can generalize across different SQL dialects, ultimately hindering the creation of robust and adaptable text-to-SQL solutions for real-world applications~\citep{spider2,li2024bird,compilersqlgen_dialect}.

\noindent\textbf{Rule-Based Translation is Insufficient.} 
% Rule-based translation provides a deterministic but rigid approach to SQL dialect conversion, struggling with complex transformations and lacking adaptability. Rule-based transpilers like SQLGlot~\citep{sqlglot} offer a structured way to convert SQLite SQL to MySQL/PostgreSQL but fail to handle complex syntax differences, such as JSON data types, strict schema constraints, and dialect-specific optimizations~\citep{zmigrod2024translating}. They also lack generalizability across databases, requiring separate rule sets for each dialect~\citep{li2024bird,spider2}. More critically, rule-based translation cannot generate diverse SQL formulations or natural language variations, limiting dataset scalability. Maintaining extensive rule-based systems is costly and impractical, as they still fail to fully capture dialect-specific nuances. Instead, an iterative AI approach ensures correctness through execution verification while expanding linguistic and structural diversity.
Rule-based translation offers a deterministic but rigid solution to SQL dialect conversion. While transpilers like SQLGlot~\citep{sqlglot} provide structured mappings between dialects, they struggle with complex syntax, schema constraints, and dialect-specific functions~\citep{zmigrod2024translating}. Moreover, these systems lack generalizability, require dialect-specific rules~\citep{li2024bird,spider2}, and cannot guarantee accurate translation. In practice, they still rely on execution-time feedback to detect and fix failures. Maintaining such rule sets is costly and brittle. Even with carefully crafted rules, such systems cannot guarantee perfect accuracy—particularly for complex or edge-case queries—and often rely on execution-time feedback for correction. We provide a detailed analysis in the Appendix~\ref{appendix:Rule_base_limitations}.

\noindent\textbf{Existing Data Collection and Training Lacks Execution Verification.} General LLM-based code data generation methods~\cite{wei2023magicoder,wang2022self} often fail to account for the specific requirements of text-to-SQL tasks, leading to the creation of syntactically plausible but incorrect SQL queries. These approaches typically generate large amounts of unverified data, which hinders their usefulness for training reliable models. Since SQL outputs can be directly validated through execution, a more structured approach that incorporates execution-based verification and targeted rejection sampling strategies is necessary. Besides, we argue that standard supervised fine-tuning (SFT) alone is insufficient to fully exploit the potential of execution validation, as it does not inherently enforce correctness across dialects.

% To address these challenges and advance dialect text-to-SQL research, we emphasize the need for a large-scale, high-quality, and executable dataset containing diverse (text, SQL) pairs. The workflow leverages LLM-based generation, iterative automatic execution validation, and self-taught error correction. In detail, the framework contains: 
To advance dialect text-to-SQL, we emphasize the importance of both high-quality, executable (text, SQL) data and a training pipeline that directly interacts with the execution environment. We propose an agentic data generation loop that combines LLM-based generation, execution-time validation, and self-correction. This offline loop yields reliable training signals, which are distilled into a dialect-aware model through supervised fine-tuning and offline reinforcement learning. The overall workflow includes:

(a) SFT Data Bootstrapping via LLM-based Translation: To mitigate the sparsity of dialect text-to-SQL data and enable effective cold-start training, we leverage high-resource SQLite (text, SQL) pairs and LLMs to efficiently sample dialect SQL queries. This bootstrapped dataset serves as a cold-start fine-tuning set, enabling rapid adaptation to low-resource dialects while minimizing manual annotation. 
%It also initializes the agentic generation loop by providing a warm-start model for dialect-aware SQL synthesis.

(b) Iterative SFT Data Generation via Execution-based Rejection Sampling: 
We extend the dataset via an iterative generation–execution–filtering loop, where the model proposes dialect SQLs executed in real databases. Valid outputs are retained through execution-aware rejection sampling, with best-of-N selection enhancing reliability. This agentic cycle uses execution feedback to govern data collection, producing higher-quality training signals without manual effort.
% Building upon the initial dataset, we incorporate small-scale dialect data and an iterative generation-execution pipeline to gradually adapt the model to dialects. Specifically, we first generate new target-dialect SQL queries referred to database, which makes it possible to be executed in the respective database environments. 

% We employ \textbf{execution-aware rejection sampling} to iteratively collect correct dialect SQL and refine the model through training on these validated samples. Successfully executed queries are retained for fine-tuning, while a \textbf{best-of-N selection strategy} with rejection finetuning prioritizes the most reliable SQL outputs. This \textbf{self-taught training process} enables the model to continuously improve its dialect SQL generation while minimizing manual supervision.  

(c) Preference Collection via Execution Feedback Rejection Sampling: To further incorporate execution feedback, we distinguish failure types and extract preference pairs—valid versus invalid SQLs—based on their execution results. These are used to train the model with DPO, which guides learning toward executable outputs. This procedure aligns with offline reinforcement learning, leveraging historical execution trajectories to improve model behavior.

We summarize our contributions as follows:
\begin{itemize}[itemsep=2pt, parsep=0pt, topsep=2pt]
\item We propose an agentic data generation loop that combines LLM-based SQL generation, execution-aware rejection sampling, and iterative self-refinement to construct high-quality dialect-specific training data with minimal manual labeling.
\item We introduce an offline reinforcement learning framework that captures execution-based preference signals and applies DPO to align the model toward generating executable SQL.
\item We conduct extensive evaluations across diverse SQL dialects (PostgreSQL, MySQL, and Oracle) $\times$ difficulty levels (single domain, cross-domain, extensive database), demonstrating significant improvements over strong baselines (e.g., GPT-4o) and providing insights for execution-guided SQL modeling.
\end{itemize}

\section{Related Work}

\subsection{Text-to-SQL}
Relational databases store a significant portion of the world's data, and retrieving information from them typically requires writing SQL queries. Automating SQL generation can lower the barrier for users to access data. A common scenario for automatic SQL generation is querying databases using natural language input~\citep{wikisql2017, yu2018spider}.  Early research treated text-to-SQL as a semantic parsing problem, where models such as RNNs and transformer-based encoders (e.g., BERT) were trained to map natural language questions to SQL statements~\citep{gan-etal-2021-natural-sql, wikisql2017,DBLP:conf/coling/Deng0022}. Performance has also improved by incorporating additional constraints into inputs and outputs~\citep{liu2021tapex, wang-etal-2021-learning-executions,deng2021spiderrealistic}. With the emergence of large language models (LLMs)~\citep{GPT3,DBLP:conf/nips/OuyangInstructGPT22,GPT4}, text-to-SQL has been further developed using prompt-based methods and fine-tuning, benefiting from LLMs' strong instruction-following and intent understanding capabilities~\citep{dong2023c3sql, li2024codes, 2023-din-sql,wang2023macsql,2024-chess}. In practical applications, text-to-SQL has been used to handle more complex data and agent-based workflows~\citep{spider2,li2024bird}.  One challenge in real-world scenarios is handling SQL dialect differences. Early studies in domain-specific languages explored this problem using intermediate meaning representations~\citep{DBLP:conf/emnlp/GuoLLLLXL20}. Some studies have attempted to address this issue through rule-based translation and compiler-based methods~\citep{compilersqlgen_dialect,lin2024momq_dialect}.  

Given the LLM-driven paradigm, this work focuses on a data-centric approach to text-to-SQL. Specifically, execution-based methods are explored to handle SQL dialect variations.

\subsection{Code LLMs}
Code foundation models have demonstrated strong code generation capabilities across various tasks. OpenAI’s Codex~\citep{codex} was one of the earliest domain-specific LLMs for coding, supporting the Copilot service~\citep{copilot}. The open-source community has further contributed with models like Deepseek-Coder~\citep{deepseek-coder} and StarCoder~\citep{li2023starcoder}, which were trained from scratch on massive code-related datasets. While others, like Code-Llama~\citep{roziere2023codellama} and Code-Qwen~\citep{hui2024codeqwen2}, adapted general-purpose models through post-training on code-specific corpora. Beyond foundation models, researchers have fine-tuned them for specific applications. Maigcoder~\citep{wei2023magicoder} enhances instruction-following abilities using curated code snippets, while Wizard-Coder~\citep{luo2023wizardcoder} and WavCoder~\citep{yu2023wavecoder} refine instruction-code alignment via evol-instruct~\citep{xu2023wizardlm}. OctoCoder~\citep{muennighoff2023octopack_OctoCoder} leverages Git commits to enhance model adaptability. Additionally, approaches like IRCoder~\citep{paul2024ircoder} and UniCoder~\citep{sun2024unicoder} explore intermediate representations (e.g., LLVM) to improve code generation. 

Compared to these approaches, our work also focuses on code generation but emphasizes leveraging execution signals from database environment. From the perspective of code LLM development, this approach provides insights applicable to broader code generation tasks. The Dialect SQL scenario serves as a practical testbed, allowing for clearer validation of method effectiveness.

\subsection{Data Synthesis}
Modern machine learning methods typically require large-scale and high-quality datasets~\citep{zhou2023lima, gao2023selfguidednoisefreedatageneration} for effective learning. However, obtaining high-quality data for every corner case is often impractical, leading researchers to explore dataset generation. By integrating existing incomplete data with the extensive knowledge embedded in LLMs, data generation can produce more comprehensive datasets for model training~\citep{DBLP:conf/acl/WangSelfInstruct23, xu2023wizardlm, wei2023magicoder}.  Recently, to enhance the reasoning capabilities of LLMs, particularly in math and code, many approaches have incorporated verifiers, such as answer or reward models, to curate high-quality datasets for model refinement~\citep{yuan2023scaling_rft_rejectionsampling,guo2025deepseekr1,zelikman2022star}.  There has also been many previous work that explores data synthesis for vision-language models~\citep{gao2023gllavasolvinggeometricproblem, pi2024personalizedvisualinstructiontuning, liu2024mitigatinghallucinationlargemultimodal, liu2024videodpoomnipreferencealignmentvideo, pi2024imagetextualizationautomaticframework, chen2024allavaharnessinggpt4vsynthesizeddata}

Our work focuses on SQL execution verification. By utilizing execution results, we obtain high-quality data by rejection sampling and further refine the model through self-taught training.

% In contrast to works that focus solely on improving code completion abilities in PLs~\citep{cassano2024knowledge}, our research aims to enhance code LLMs' instruction-following capability for LRPLs. We achieve this by improving the alignment between NL and PL during the fine-tuning stage.

\section{Methodology}
In this section, we present the details of our approach to obtain \method, including 3 phases: Translation Bootstrapping, Iterative Data Generation and Training, and Preference Enhancement. The key idea of Execution-Assisted Generation is fully leveraging execution verification signals to asisst LLM to generate high-quality data for text-to-SQL across different dialects. An illustration of \method ~is shown in Figure~\ref{fig:iterative_training}.

\subsection{Formulation}

% To formally describe our approach, we define a natural language question as \( Q \) and its corresponding SQL query as \( S \). A dataset of (question, SQL) pairs is denoted as $D = \{(Q_i, S_i)\}_{i=1}^{N}$.

% Given that most existing datasets are written in a single SQL dialect, we define \( D_{\text{Source}} \) as an existing dataset with SQL queries in a source dialect (e.g., SQLite) and \( D_{\text{Target}} \) as the translated dataset with queries adapted to target dialects (e.g., MySQL, PostgreSQL). To perform this translation, we introduce a function, $T: S_{\text{Source}} \to S_{\text{Target}}$ that converts SQL from one dialect to another.

% Our model, denoted as \( M_{\theta} \), is an LLM parameterized by \( \theta \), trained to generate SQL queries given a natural language question. Since generated SQL queries may contain errors, we incorporate an execution verification function, \( R(S) \), to check whether an SQL query \( S \) executes successfully.

% To improve SQL quality, we apply an iterative refinement process based on execution results. Specifically, we collect incorrect queries into a negative dataset \( D_{\text{Neg}} \), allowing the model to learn from execution failures. Through an iterative process of translation, execution validation, and contrastive learning using failure cases, our approach enhances the robustness of SQL generation across multiple dialects while constructing a high-quality multi-dialect text-to-SQL dataset.

We denote a natural language query as \( Q \), its corresponding SQL as \( S \), and the generation model as an LLM \( M_\theta \). The training set \( \mathcal{D} = \{(Q_i, S_i)\}_{i=1}^{N} \) is constructed by translating a high-resource source dialect \( \mathcal{D}_{\text{Source}} \) (e.g., SQLite) to target dialects using a bootstrapping model and a dialect mapping function \( T \).

To guide model training, we define an execution-based reward function \( \mathcal{R}(S) \in \{0,1\} \), which returns 1 if the SQL executes successfully. The goal is to train a model that maximizes expected execution success:

\begin{equation}
\pi_\theta^* = \arg\max_{\pi_\theta} \; \mathbb{E}_{Q \sim \mathcal{D}} \left[ \mathbb{E}_{\hat{S} \sim \pi_\theta(\cdot|Q)} \left[ \mathcal{R}(\hat{S}) \right] \right]
\end{equation}

We adopt a self-evolving offline training strategy~\citep{zelikman2022star,DBLP:journals/corr/raft,DBLP:journals/corr/rest,schulman2017ppo}, which iteratively (1) filters generated SQLs via \textbf{execution-guided rejection sampling}, and (2) applies \textbf{preference optimization} through Direct Preference Optimization (DPO). The model is updated at iteration \( t \) as:

\begin{equation}
\pi_\theta^{(t+1)} = \arg\max_{\pi_\theta} \; \mathbb{E}_{Q, \hat{S}, S^* \sim \mathcal{D}} \left[ \mathcal{R}(S^*, \hat{S}) \right]
\end{equation}

Here, \( S^* \) denotes a preferred (e.g., executable) SQL, contrasted against a failed candidate \( \hat{S} \). This defines an offline reinforcement learning loop grounded in execution feedback.

\begin{wrapfigure}{r}{0.48\textwidth}
  \vspace{-4mm}
    \includegraphics[width=0.48\textwidth]{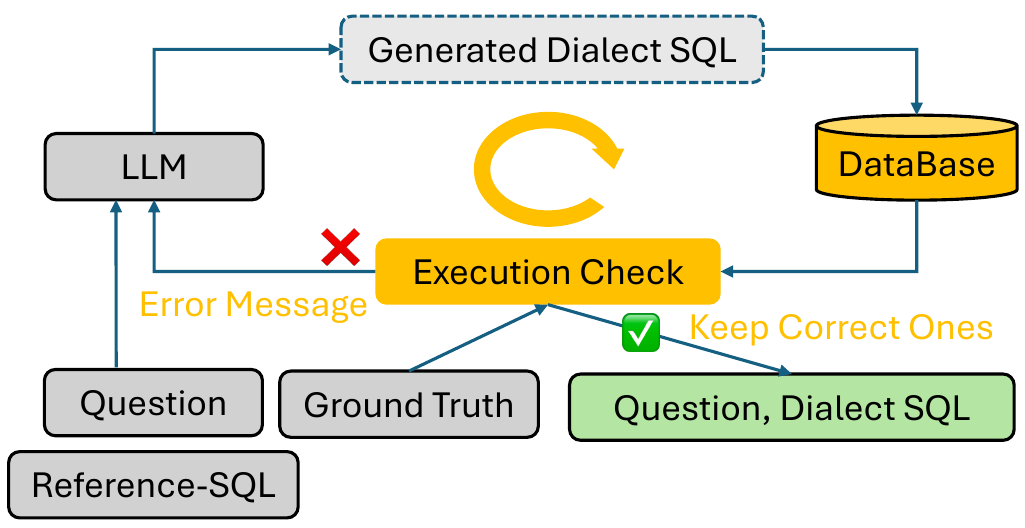}
    \caption{Execution-based error feedback loop for dialect-specific SQL refinement. Through this, we can collect a bootstrap dataset to resolve the cold-start issue of training expert dialect model.}
    \label{fig:execution_check}
  \vspace{-6mm}
\end{wrapfigure}

% \begin{wrapfigure}{r}{0.48\textwidth}
%   \vspace{-4mm}
%   \centering
%   \includegraphics[width=0.95\linewidth]{figures/new_motivation.pdf}
%   \vspace{-3mm}
%   \caption{
%     Given a natural language question, different SQL dialects require distinct syntax adjustments, such as explicit type casting in PostgreSQL. Beyond the traditional text-input–SQL-output formulation, we incorporate the database environment to enable agentic execution feedback for data synthesis and training.
%   }
%   \label{fig:motivation}
%   \vspace{-6mm}
% \end{wrapfigure}

\subsection{Translation-based Bootstrapping}

Let \( D_{\text{SQLite}} = \{ (Q_i, S_i) \}_{i=1}^{N} \) be a large-scale dataset containing natural language questions \( Q_i \) paired with corresponding SQL queries \( S_i \) written in SQLite dialect. Given the scarcity of multi-dialect SQL datasets, we first leverage \( D_{\text{SQLite}} \) to bootstrap an initial dataset for training.

To achieve this, we introduce a translation function \( T: S_{\text{SQLite}} \to S_{\text{Target}} \), which generates an SQL query \( S_{\text{Target}} \) in the target dialect based on both the original SQL query \( S_{\text{SQLite}} \) and the corresponding question \( Q \), modeled as:

\[
S_{\text{Target}} \sim P(S_{\text{Target}} | Q, S_{\text{SQLite}})
\]

However, direct translation does not guarantee correctness due to differences in SQL syntax and execution semantics across dialects. To refine the generated SQL queries, we incorporate an \textbf{execution-based verification and iterative correction mechanism}, as illustrated in Figure~\ref{fig:execution_check}.

\begin{figure*}[h]
    \centering
    \includegraphics[width=\textwidth]{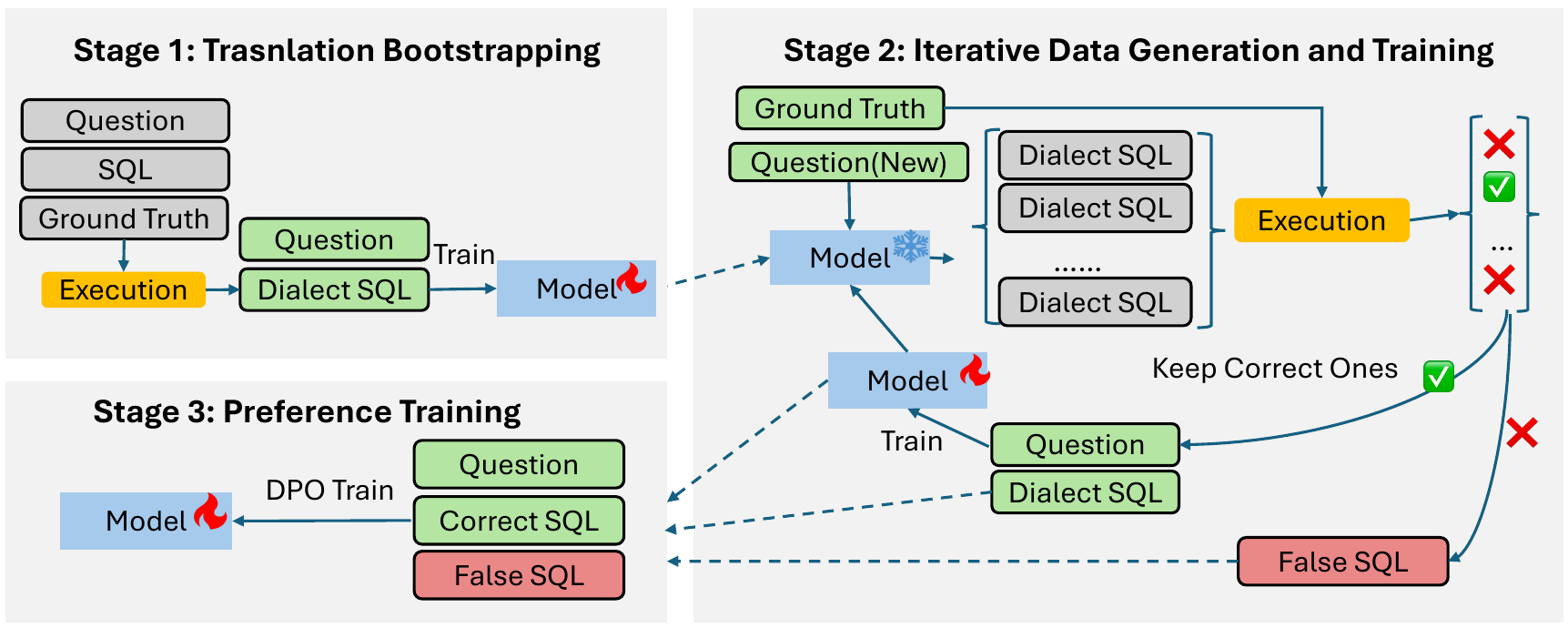}
    \caption{Pipeline for Dialect Text-to-SQL Data Generation and Model Training. The framework consists of three stages: (1) \textbf{Translation Bootstrapping}: A bootstrap text-to-SQL model is fine-tuned using SQL translations from an existing dataset (e.g., SQLite) to other dialects (e.g., MySQL, PostgreSQL). (2) \textbf{Iterative Data Generation and Training}: The model generates multiple SQL candidates per question, which are validated via execution feedback. Correct queries are retained to refine the dataset, enabling iterative self-improvement. (3) \textbf{Preference Enhancement}: A Direct Preference Optimization (DPO) step is applied to distinguish correct and incorrect SQL queries. High-quality pairs (question, correct SQL) are used to further improve the model’s performance and preference learning, ensuring both correctness and efficiency in SQL generation.}
    \label{fig:iterative_training}
\end{figure*}

The refinement process operates as follows (Appendix~\ref{appendix:translation}): 1) An LLM (GPT-4o here) generates candidate SQL queries \( S_{\text{Target}} \) for a given natural language question \( Q \), conditioned on \( S_{\text{SQLite}} \).
2) The generated SQL query is executed in a database corresponding to the target dialect.
3) If the execution succeeds, the query is added to the validated dataset:
   $D_{\text{Trans}} = \{(Q_i, S_{\text{Target},i})\}$
4) If the execution fails, the database returns an error message, which is fed back into the LLM as an additional context for refining the SQL query. The model iteratively refines \( S_{\text{Target}} \) until a valid query is produced.
5) This iterative execution check continues until either a valid SQL query is found or a maximum refinement threshold is reached.

This approach effectively corrects syntactic and semantic errors by leveraging real execution feedback rather than relying solely on static rule-based translation. Through this execution-aware iteration, the model progressively learns to generate more accurate and dialect-specific SQL queries. The final dataset, \( D_{\text{Trans}} \), serves as a high-quality dialect training corpus, enabling robust generalization across different database systems.

\subsection{Iterative Data Generation and Training 
}

While \( D_{\text{Trans}} \) provides a baseline, rule-based translation alone is insufficient to guarantee correctness due to syntax differences, type constraints, and execution behaviors across SQL dialects. To address this, we introduce an iterative execution-feedback process incorporating rejection sampling and augmented question generation, as depicted in Figure~\ref{fig:iterative_training}.

\subsubsection{Augmenting Training Data with New Questions}

To improve model generalization across SQL dialects, we incorporate additional natural language questions from two sources:\textbf{(1) Existing Text-to-SQL Datasets}:  
    We extract additional questions from existing datasets like WikiSQL, ensuring coverage of diverse query structures. \textbf{(2) Database-Aware Question Generation}:  
    We leverage \textbf{GPT-4o} to generate new questions based on actual database values.  
    Given a schema and sample database records, GPT-4o generates contextually relevant questions that reference specific values, improving the model’s robustness in handling real-world queries.

By integrating these new questions, we expand our dataset beyond simple rule-based translations, allowing the model to generate and validate SQL queries for a more diverse set of inputs.

\subsubsection{Execution-based Rejection Sampling}

For each natural language question \( Q_i \), the model \( M_{\theta} \) generates multiple dialect-specific SQL candidates \( \{ S_{\text{cand},i} \} \), following the probability distribution:$S_{\text{cand},i} \sim P_{\theta}(S | Q_i)$

Each candidate query is then executed in the corresponding database environment, yielding an execution result \( R(S_{\text{cand},i}) \):
\[
R(S) =
\begin{cases} 
1, & \text{if } S \text{ executes successfully} \\
0, & \text{if } S \text{ fails due to execution errors}
\end{cases}
\]
We apply a rejection sampling to iteratively refine SQL generation: If $S_{\text{cand}}$ exectues successfully, i.e., $R(S_{\text{cand},i}) = 1 $. The query is added to the validated dataset:
      $D_{\text{Valid}} = D_{\text{Valid}} \cup \{ (Q_i, S_{\text{cand},i}) \}$
    
If $S_{\text{cand}}$ is a Failure Case, i.e., $ R(S_{\text{cand},i}) = 0$. The query is stored in the negative dataset:
      $
      D_{\text{Neg}} = D_{\text{Neg}} \cup \{ (Q_i, S_{\text{cand},i}) \}
      $

This process is iteratively repeated until a valid SQL query is generated or a predefined iteration limit is reached.

\subsubsection{Iterative Data Generation and Model Refinement}

The validated dataset \( D_{\text{Valid}} \) is used for further fine-tuning, while incorrect queries in \( D_{\text{Neg}} \) serve as contrastive learning signals in later preference optimization stages. 

This process results in a high-quality, dialect-aware text-to-SQL dataset that is continuously refined through execution-based validation and real-world query augmentation.

\subsection{Preference Optimization}

To further refine the model’s SQL generation capabilities, we leverage DPO~\citep{rafailov2023direct} to distinguish between correct and incorrect queries, using execution feedback as the primary signal. The negative dataset \( D_{\text{Neg}} \) and validated dataset \( D_{\text{Valid}} \) have already been collected during the Iterative Data Generation and Training phase. Here, we construct preference pairs to fine-tune the model based on execution outcomes.

\paragraph{Pairwise Preference Data Construction}

To enable preference learning, we form query pairs \( (S_{\text{pos}}, S_{\text{neg}}) \), where: $S_{\text{pos}} \in D_{\text{Valid}}, \quad S_{\text{neg}} \in D_{\text{Neg}}$
These pairs allow the model to differentiate between correct and incorrect SQL, ensuring that preference learning reinforces correct generation.

\paragraph{Direct Preference Optimization (DPO) Training}

The model is fine-tuned using DPO, where the objective is to maximize the probability of generating preferred SQL queries over non-preferred ones:

\[
P_{\theta}(S_{\text{pos}} | Q) > P_{\theta}(S_{\text{neg}} | Q)
\]

By leveraging execution failures as negative examples and correct executions as positive examples, the model learns to generate more reliable and executable SQL queries. This approach enhances both the correctness and robustness of SQL generation across different dialects.

\section{Implementation and Evaluation Settings}

The bootstrap dataset and new questions for \method~are generated using GPT-4o~\citep{GPT4}. We choose GPT-4o due to its superior ability to follow instructions and leverage error messages to generate accurate bootstrap dialect SQL examples. The final \method~dataset consists of 20.6k  samples in the supervised finetuning (SFT) set and 8k samples in the preference pairs (Appendix~\ref{appendix:generated_data}).
All training is conducted on four A6000 GPUs. We fine-tune the full-parameter Deepseek-Coder-7B~\citep{deepseek-coder} for supervised finetuning (SFT) and Direct Preference Optimization (DPO). For detailed training configurations and inference hyperparameters, please refer to Appendix~\ref{appendix:implementation}

For baseline comparisons, we evaluate GPT-4o-2024-11-20 and Gemini-1.5-pro-0827~\citep{reid2024gemini}, both of which were released in 2024. Since these models were trained on publicly available data up to their release dates, they likely include extensive SQL-related training data, ensuring a fair comparison. 

\subsection{Text-to-SQL across dialects and Benchmarks}
\paragraph{Dialects.} To fully validate the generalization ability of our method, we selected three SQL dialects: \texttt{PostgreSQL}, \texttt{MySQL} and \texttt{Oracle}. Our pipeline is dialect-agnostic, we chose these two dialects to verify the generalizable effectiveness of our pipeline across different dialects.

\begin{table*}[ht]
    \centering
    \renewcommand{\arraystretch}{1.2}
    \small % Reduce font size for the entire table environment
    \setlength{\tabcolsep}{6pt} % Reduce space between columns (default is ~6pt)

    % Tabular definition with 9 columns (l c cc ccc c c)
    \begin{tabular}{l c cc ccc c c}
        \toprule
        % Header rows - MySQL now spans 3 columns
        \multirow{2}{*}{\textbf{Method}} & \multirow{2}{*}{\textbf{Model size}} & \multicolumn{2}{c}{\textbf{PostgreSQL}} & \multicolumn{3}{c}{\textbf{MySQL}} & \multicolumn{1}{c}{\textbf{Oracle}} & \multirow{2}{*}{\textbf{Average}} \\
        \cmidrule(lr){3-4} \cmidrule(lr){5-7} 
        % Sub-headers including 'Bird' under MySQL
        &  & Spider & WikiSQL & Spider & WikiSQL & Bird & Spider &  \\
        \midrule

        % Unfinetuned LLM Section - Corrected multicolumn span to 9
        \multicolumn{9}{l}{\textit{General purposed LLM}} \\
        \textbf{GPT-4o} & - & 54.59 & 58.97 & 62.09 & 57.24 & 36.38 & 64.86 & 55.69 \\
        \textbf{Gemini-1.5-pro} & - & 51.03 & 54.1 & 64.90 & 51.95 & 36.11 & 65.21 & 53.88 \\
        %\textbf{Deepseek-V3} & 671B & 27.99 & 17.56 & 26.55 & 21.51 & - & - & 23.40 \\
        \textbf{Llama3.1-Instruct} & 8B & 33.63 & 31.6 & 48.86 & 25.41 & 24.58 & 30.0 & 32.35 \\

        \midrule
        % Code Expert LLM Section - Corrected multicolumn span to 9
        \multicolumn{9}{l}{\textit{Code Expert LLM}} \\
        \textbf{Deepseek-Coder} & 7B & 37.31 & 18.12 & 49.6 & 24.67 & 16.00 & 50.77 & 32.75 \\
        \textbf{Qwen-Coder} & 7B & 36.8 & 15.48 & 39.04 & 22.84 & 15.36 & 58.31 & 31.31 \\
        \textbf{Magicoder} & 7B & 21.9 & 17.45 & 47.28 & 23.32 & 13.23 & 26.6 & 24.96 \\
        \textbf{WizardCoder} & 15B & 23.78 & 16.91 & 32.36 & 20.56 & 18.38 & 36.33 & 24.72 \\

        \midrule
        % SQL Expert LLM Section - Corrected multicolumn span to 9
        \multicolumn{9}{l}{\textit{SQL Expert LLM}} \\
        \textbf{CodeS} & 7B & 24.76 & 20.0 & 35.6 & 23.0 & 14.41 & 37.4 & 25.86 \\
        \textbf{StructLLM} & 7B & 38.71 & 30.97 & 44.2 & 7.14 & 22.69 & 33.16 & 29.48 \\

        \midrule
        % ExeSQL Row (Last method)
        \textbf{ExeSQL} & 7B & \textbf{69.86} & \textbf{74.10} & \textbf{72.09} & \textbf{73.64} & \textbf{41.13} & \textbf{69.35} & \textbf{66.70} \\
        \bottomrule
        
    \end{tabular}
    \caption{Performance comparison of various LLMs on Dialect text-to-SQL benchmarks. \method ~surpasses all baseline models, achieving an average improvement of 11.0\% over GPT-4o.}
    \label{tab:main_sql_llm_comparison} % Updated label slightly
\end{table*}

\paragraph{Benchmarks.}
We adapt three standard benchmarks, Spider~\citep{yu2018spider} WikiSQL~\citep{wikisql2017} and BIRD~\citep{li2024bird}, for in-domain evaluation and use Dr.Spider~\citep{chang2023dr} as an out-of-domain dataset. We also incorporate the single-domain benchmark MimicSQL~\citep{wang2020texttosqlgenerationquestionanswering, deng-etal-2022-recent} to evaluate our model across varying difficulty levels. For dialect SQL evaluation, we extract the question, database, and ground truth result, prompting the model to generate dialect-specific SQL and verifying execution accuracy. Details on these datasets are in Appendix~\ref{appendix:eval_data}. To ensure accurate evaluation, we preprocess responses to extract SQL using an answer extraction tool (Appendix~\ref{appendix:answer_extraction}). For results on the single-domain dataset, please refer to Appendix~\ref{sec:single_domain_generalization}.

\subsection{Baseline Models}  
\textbf{General purposed LLM baselines:} We evaluate four large language models (LLMs) without any fine-tuning for text-to-SQL tasks: \texttt{GPT-4o}~\citep{GPT4}, \texttt{Gemini-1.5-pro}~\citep{reid2024gemini}, %\texttt{Deepseek-V3}~\citep{liu2024deepseekv3},
and \texttt{Llama3.1-Instruct}~\cite{metallama3}. These models are assessed by directly prompting them to generate SQL queries given a natural language question and the corresponding database schema.

\noindent\textbf{Code Expert LLM baselines:} These baselines consist of LLMs trained on large-scale code-related corpora, making them well-suited for code generation tasks. We include \texttt{DeepSeek-Coder}~\citep{deepseek-coder}, \texttt{Qwen-Coder}~\citep{hui2024codeqwen2}, \texttt{Magicoder-DS}~\citep{wei2023magicoder}, and \texttt{WizardCoder}~\citep{luo2023wizardcoder}.

\noindent\textbf{SQL Expert LLM baselines:} Several LLMs are specifically adapted for SQL generation, typically optimized for the SQLite dialect and demonstrating strong table understanding capabilities. We include \texttt{Code-S}~\citep{li2024codes} and \texttt{StructLLM}~\citep{zhuang2024structlm} in this category. 

\noindent The comparisons in (2) and (3) aim to assess whether fine-tuned general-purpose LLMs can outperform specialized code-generation or SQL-focused models in specific scenarios.

\section{Experimental Results}

\subsection{Main Results}

We present the main experimental results in Table~\ref{tab:main_sql_llm_comparison}. From the table, we observe that \method~ achieves an average accuracy of 66.70\% across PostgreSQL, MySQL and Oracle benchmarks, significantly outperforming all baseline models.  

\textbf{General purposed LLMs.} Among the general-purpose LLMs, GPT-4o achieves the highest accuracy (55.69\%), demonstrating its strong zero-shot SQL generation capability. We find that Gemini-1.5-pro underperforms GPT-4o, achieving 53.88\%. Llama3.1-8B-Instruct perform worse, with average accuracies of 32.35\%, respectively. These results indicate that general-purpose LLMs struggle with SQL dialect variations.  

\textbf{Code Expert LLMs.} Code-focused models, such as Deepseek-Coder and Qwen-Coder, demonstrate better performance than standard LLMs. Deepseek-Coder achieves an average accuracy of 32.75\%, while Qwen-Coder reaches 31.31\%. However, Magicoder and WizardCoder perform worse, suggesting that general code generation ability does not equal SQL generation (especially dialect) capability. This implies that code training alone is insufficient for SQL dialect adaptation.  

\textbf{SQL Expert LLMs.} The SQL-specialized models exhibit the most significant improvements. StructLLM, which is trained on SQL-specific tasks, achieves an accuracy of 29.48\%, slightly outperforming most code models. However, \method~ surpasses all baselines by a large margin, reaching an average accuracy of 66.70\%. Also, it is worth noting that these models often have a great performance degradation compared with SQLite performance (Appendix~\ref{appendix:dialect_degradation}).

These results highlight the importance of the proposed execution-based fine-tuning and dialect-aware SQL adaptation. Unlike general-purpose or code-focused models, \method~ effectively learns to handle different SQL dialects through iterative refinement, leading to a substantial performance boost. 
%The significant gap between \method~ and baseline models suggests that current LLMs, even code-trained ones, lack sufficient SQL dialect understanding, emphasizing the necessity of specialized fine-tuning for text-to-SQL tasks.

\subsection{Further Analysis}
To validate the effectiveness of \method, we conduct three analyses: (1) Ablation studies assess the impact of iterative refinement and preference learning on accuracy. (2) ID and OOD evaluation measures generalization to unseen queries and SQL dialects. (3) Execution-based rejection sampling analysis examines its role in improving SQL correctness. These analyses confirm \method’s robustness and adaptability.

\begin{table}[ht]
\centering
\begin{minipage}[t]{0.47\textwidth}
    \centering
    \renewcommand{\arraystretch}{1.0}
    \setlength{\extrarowheight}{2pt}
    \begin{tabular}{lcc}
        \toprule
        \textbf{Method} & \textbf{PostgreSQL} & \textbf{MySQL} \\
        \midrule
        ExeSQL & 71.98 & 72.87 \\
        w/o iteration & 63.49 & 60.09 \\
        w/o preference & 71.36 & 70.34 \\
        \bottomrule
    \end{tabular}
    \vskip 0.1in
    \caption{Performance comparison of different ExeSQL ablations.}
    \label{tab:exeSQL_ablations}
\end{minipage}
\hfill
\begin{minipage}[t]{0.49\textwidth}
    \centering
    \renewcommand{\arraystretch}{1.0}
    \resizebox{\linewidth}{!}{
    \begin{tabular}{c cc cc}
        \toprule
        \multirow{2}{*}{\textbf{Method}} & \multicolumn{2}{c}{\textbf{PostgreSQL}} & \multicolumn{2}{c}{\textbf{MySQL}} \\
        \cmidrule(lr){2-3} \cmidrule(lr){4-5}
        & \textcolor{turquoisegreen}{Spider} & \textcolor{babyblueeyes}{Dr.} & \textcolor{turquoisegreen}{Spider} & \textcolor{babyblueeyes}{Dr.}  \\
        \midrule
        Deepseek-Coder & 37.31 & 27.10 & 49.60 & 36.82 \\
        StructLLM & 38.71 & 25.83 & 44.20 & 40.00 \\
        ExeSQL & \textbf{69.86} & \textbf{59.16} & \textbf{72.09} & \textbf{56.02} \\
        \bottomrule
    \end{tabular}
    }
    \caption{Results on \textcolor{turquoisegreen}{ID} and \textcolor{babyblueeyes}{OOD} evaluation. \method~shows strong generalization without overfitting.}
    \label{tab:sql_id_ood}
\end{minipage}
\end{table}

\vspace{0.5em}
\subsubsection{Ablations for Iteration Data Generation}
\label{analysis: ablation}
Table~\ref{tab:exeSQL_ablations} shows that removing iteration-based refinement significantly reduces performance (\textbf{71.98\%} to \textbf{63.49\%} on PostgreSQL, \textbf{72.865\%} to \textbf{60.09\%} on MySQL), highlighting the importance of iterative data generation in improving SQL accuracy. Removing preference learning also leads to a performance drop, though less severe, indicating that preference optimization further refines query quality. These results demonstrate that both iterative refinement and preference learning play crucial roles in enhancing \method’s effectiveness.

\subsubsection{ID and OOD Evaluation.}
\label{analysis: ood}
\begin{wrapfigure}{r}{0.48\textwidth}
\vspace{-3em}
    \includegraphics[width=0.5\textwidth]{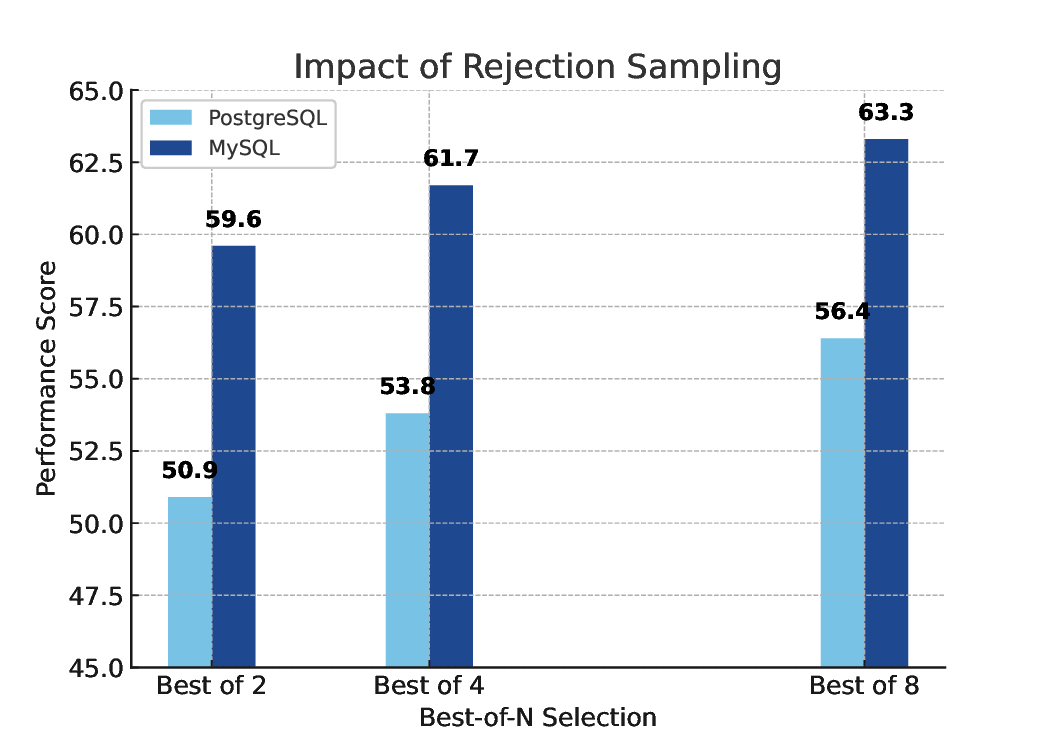}
    \vspace{-2em}
    \caption{Retention rate of correct dialect SQL under different best-of-N sampling strategies on 1,000 queries. Results show the bootstrapped model already produces many correct samples, with larger N further improving correctness.}
    \label{fig:rejection_sampling}
    \vspace{-3em}
\end{wrapfigure}
We evaluate \method~on both in-distribution (ID) and out-of-distribution (OOD) datasets to assess its generalization. The OOD evaluation is conducted on \textbf{Dr.Spider}~\citep{chang2023dr}, a diagnostic text-to-SQL benchmark with 15,269 samples, introducing perturbations in databases (DB), natural language queries (NLQ), and SQL to test robustness. Given its scale, Dr.Spider is significantly harder to overfit than Spider’s 2,147 samples.  

Table~\ref{tab:sql_id_ood} shows that \method~consistently achieves the highest accuracy across all settings. Notably, ExeSQL outperforms StructLLM and Deepseek-Coder by a large margin on both PostgreSQL and MySQL, confirming its strong generalization to both ID and OOD queries.

% \begin{figure}[ht]
%     \centering
%     \includegraphics[width=0.5\textwidth]{figures/best_of_n_performance.eps}
%     \caption{Retention rate of correct dialect SQL samples across different best-of-N sampling strategies, evaluated on 1,000 query samples. The results indicate that the bootstrapped model can already generate many correct dialect samples, and increasing N further improves correctness.}
%     \label{fig:rejection_sampling}
% \end{figure}

\subsubsection{Configuration of Execution-based Rejection Sampling.}
\label{analysis: best-of-n}

Figure~\ref{fig:rejection_sampling} presents the effect of execution-based rejection sampling on SQL generation accuracy across different best-of-N selection strategies. As $N$ increases, the proportion of correct dialect SQL samples improves consistently for both PostgreSQL and MySQL.

This result indicates that the bootstrapped model is capable of generating a significant number of correct dialect SQL queries even without additional fine-tuning. The primary challenge then shifts to efficiently identifying and selecting these correct samples. An iterative sampling approach can be employed to extract high-quality SQL queries, which can further enhance the model through self-supervised training.

\section{Conclusion}

We propose an execution-driven framework to enhance text-to-SQL generation across multiple SQL dialects. By integrating LLM-based dialect bootstrapping, {execution feedback rejection sampling, and preference learning, our approach iteratively refines SQL generation through execution validation and error correction. Experiments show that \method~ outperforms GPT-4o by a large margin on 3 dialects, respectively, demonstrating superior adaptability and correctness. Our findings highlight the importance of execution-aware training and provide a scalable solution for robust multi-dialect text-to-SQL modeling.

\section*{Limitations}
In this work, we primarily focus on two mainstream dialects (MySQL and PostgreSQL) within a relatively simple environment. This setting overlooks complexities that arise in larger-scale or heterogeneous scenarios, and it only partially addresses advanced dialect-specific features (e.g., complex window functions or Regex handling). Moreover, our iterative generation process relies on predefined prompts and partial rules, which may not readily accommodate databases with significantly different formal grammars. In future research, we plan to explore more dialects and more complex database conditions, aiming to enhance the coverage and robustness of our multi-dialect text-to-SQL framework.

% \section*{Ethics Statement}
% Scientific work published at ACL 2023 must comply with the ACL Ethics Policy.\footnote{\url{https://www.aclweb.org/portal/content/acl-code-ethics}} We encourage all authors to include an explicit ethics statement on the broader impact of the work, or other ethical considerations after the conclusion but before the references. The ethics statement will not count toward the page limit (8 pages for long, 4 pages for short papers).

\input{template.bbl}

\bibliographystyle{unsrtnat}
\bibliography{template}

\appendix
\section{Appendix}
\subsection{Analysis for the Dialect-Degradation for Text-to-SQL }
\label{appendix:dialect_degradation}

\begin{table}[ht]
    \centering
    \renewcommand{\arraystretch}{1.2}
    \setlength{\tabcolsep}{8pt}
    \begin{tabular}{lccc|c}
        \toprule
        \textbf{Method} & \textbf{PostgreSQL } & \textbf{MySQL } & \textbf{SQLite } & $\Delta$ \\
        \midrule
        GPT-4o & 54.59 & 62.09 & 71.17 & -12.83 \\
        Gemini-1.5-pro & 51.03 & 64.90 & 77.27 & -19.31 \\
        CodeS & 24.76 & 35.60 & 77.90 & -47.72 \\
        StructLLM & 38.71 & 44.20 & 70.20 & -28.75 \\
        \bottomrule
    \end{tabular}
    \vskip 0.05in
    \caption{Zero-shot performance comparison on Spider across different SQL dialects. Flip $\Delta$ measures the differences between the model's performance on SQlite compared with PostgreSQL and MySQL. }
    \label{tab:sql_dialect_comparison}
\end{table}

Table~\ref{tab:sql_dialect_comparison} demonstrates that general LLMs experience significant performance degradation across SQL dialects, with Flip $\Delta$ ranging from -12.83 to -47.72. Larger models such as GPT-4o and Gemini-1.5-pro degrade less, while smaller models like CodeS and StructLLM suffer more. In contrast, SQL-Expert models exhibit even 2-3$\times$ higher degradation, likely due to weaker generalization from smaller parameter sizes. This highlights the importance of SQL dialect adaptation research, as even strong general LLMs struggle with dialect shifts.

\subsection{Generated Data Statistics}
\label{appendix:generated_data}

Table~\ref{tab:dataset_stats} shows the distribution of question sources. During SFT data generation, 3.7k new dialect-specific questions were created based given the values in databases.

\begin{table}[h]
    \centering
    \renewcommand{\arraystretch}{1.2}
    \begin{tabular}{lcc}
        \hline
        Stage & Dataset & Size \\
        \hline
        \multirow{3}{*}{SFT} & Spider & 6.9k  \\
                             & WikiSQL & 10k \\
                             & New Generated Data & 3.7k  \\        
        \hline
        \multirow{2}{*}{DPO} & Spider & 4k  \\
                             & WikiSQL & 4k  \\
        \hline
    \end{tabular}
    \vskip 0.06in
    \caption{Dataset statistics for SFT and DPO training stages.}
    \label{tab:dataset_stats}
\end{table}

\subsection{Implementation}
\label{appendix:implementation}
We fine-tune the full-parameter Deepseek-Coder-7B~\citep{deepseek-coder} using supervised finetuning (SFT) for one epoch with a batch size of 16 and a learning rate of 2e-5. For Direct Preference Optimization (DPO) training, we train for three epochs with a batch size of 16 and a learning rate of 5.0e-6. Additionally, we incorporate the SFT loss into the DPO loss with a weight of 1 during preference training.
For execution-based rejection sampling and worst-of-n negative sample collection, we set the inference parameters to temperature = 0.7, top-p = 0.9, and top-k = 50. Negative examples for DPO training are selected using the worst-of-N strategy, with N = 8.

Here, we make use of huggingface-transformer~\citep{wolf-etal-2020-transformers} and llama-factory~\citep{zheng2024llamafactory} to perform training. For the inference, we make use of vllm toolkit~\citep{kwon2023vllm}.

% \begin{table}[ht]
%     \centering
%     \renewcommand{\arraystretch}{1.2}
%     \setlength{\tabcolsep}{10pt}
%     \label{tab:sft_config}
%     \begin{tabular}{lc}
%         \toprule
%         \textbf{Setting} & \textbf{Value} \\
%         \midrule
%         GPUs Used & 4 × A6000 \\
%         Model & Deepseek-Coder-7B \\
%         Batch Size & 16 \\
%         Epochs & 1 \\
%         Learning Rate & $2 \times 10^{-5}$ \\
%         \bottomrule
%     \end{tabular}
%     \vskip 0.06in
%     \caption{Supervised Fine-tuning (SFT) Configuration}
% \end{table}

% \begin{table}[ht]
%     \centering
%     \renewcommand{\arraystretch}{1.2}
%     \setlength{\tabcolsep}{10pt}
    
%     \label{tab:dpo_config}
%     \begin{tabular}{lc}
%         \toprule
%         \textbf{Setting} & \textbf{Value} \\
%         \midrule
%         GPUs Used & 4 × A6000 \\
%         Model & Deepseek-Coder-7B \\
%         Batch Size & 16 \\
%         Epochs & 3 \\
%         Learning Rate & $5 \times 10^{-6}$ \\
%         Loss Weight & 1 (SFT + DPO) \\
%         Worst-of-N & 8 \\
%         \bottomrule
%     \end{tabular}
%     \vskip 0.06in
%     \caption{Direct Preference Optimization (DPO) Configuration}
% \end{table}

\begin{table}[ht] % Use table* for full page width if needed
\centering
\label{tab:training_configs} % A single label for the combined table
\begin{minipage}[t]{0.48\textwidth} % Minipage for the first table (SFT)
\centering
\renewcommand{\arraystretch}{1.2}
\setlength{\tabcolsep}{8pt} % Slightly adjust colsep for better fit
\begin{tabular}{lc}
\toprule
\textbf{Setting} & \textbf{Value} \\
\midrule
GPUs Used & 4 $\times$ A6000 \\
Model & Deepseek-Coder-7B \\
Batch Size & 16 \\
Epochs & 1 \\
Learning Rate & $2 \times 10^{-5}$ \\
\bottomrule
\end{tabular}
\vskip 0.06in
\caption{Supervised Fine-tuning (SFT) Configuration} % Use caption* for sub-caption without number
\label{tab:sft_config} % Keep individual label for potential specific reference
\end{minipage}
\hfill % Adds horizontal space between minipages, pushing them apart
\begin{minipage}[t]{0.48\textwidth} % Minipage for the second table (DPO)
\centering
\renewcommand{\arraystretch}{1.2}
\setlength{\tabcolsep}{8pt} % Slightly adjust colsep for better fit
\begin{tabular}{lc}
\toprule
\textbf{Setting} & \textbf{Value} \\
\midrule
GPUs Used & 4 $\times$ A6000 \\
Model & Deepseek-Coder-7B \\
Batch Size & 16 \\
Epochs & 3 \\
Learning Rate & $5 \times 10^{-6}$ \\
Loss Weight & 1 (SFT + DPO) \\
Worst-of-N & 8 \\
\bottomrule
\end{tabular}
\vskip 0.06in
\caption{Direct Preference Optimization (DPO) Configuration} % Use caption* for sub-caption without number
\label{tab:dpo_config} % Keep individual label for potential specific reference
\end{minipage}
\end{table}

\begin{table}[ht]
    \centering
    \renewcommand{\arraystretch}{1.2}
    \setlength{\tabcolsep}{10pt}
    \label{tab:inference_config}
    \begin{tabular}{lc}
        \toprule
        \textbf{Setting} & \textbf{Value} \\
        \midrule
        Temperature & 0.7 \\
        Top-p & 0.9 \\
        Top-k & 50 \\
        \bottomrule
    \end{tabular}
    \vskip 0.06in
    \caption{Inference and Rejection Sampling Configuration}
\end{table}
\subsection{Impact of Direct Preference Optimization (DPO)}
\label{sec:impact_of_dpo}

To understand the performance impact of Direct Preference Optimization (DPO) following Supervised Fine-tuning (SFT), we evaluated the performance on both PostgreSQL and MySQL dialects. The results, based on an initial SFT with 8,000 samples and a subsequent DPO with 20,000 preference samples, are presented in Table \ref{tab:sft_vs_dpo}.

\begin{table}[ht]
\centering
\renewcommand{\arraystretch}{1.2}
\begin{tabular}{lccc}
\hline
Model & PostgreSQL & MySQL & Oracle \\
\hline
SFT & 71.36 & 70.34 & 65.86\\
SFT + DPO & 71.98 & 72.86 & 69.35\\
\hline
\end{tabular}
\vskip 0.06in
\caption{Performance Comparison: SFT vs. SFT + DPO}
\label{tab:sft_vs_dpo}
\end{table}

Furthermore, to analyze the detailed effect of DPO on model robustness and generalization, we evaluated both models' performance under database perturbation and SQL perturbation. The results, presented as the average performance across PostgreSQL and MySQL, are shown in Table \ref{tab:dpo_robustness}.

\begin{table}[ht]
\centering
\renewcommand{\arraystretch}{1.2}
\begin{tabular}{lccc}
\hline
Model & Database Perturbation & SQL Perturbation & Average \\
\hline
SFT & 55.54 & 62.16 & 58.85 \\
SFT + DPO & 57.14 & 64.06 & 60.60 \\
\hline
\end{tabular}
\vskip 0.06in
\caption{Robustness and Generalization Analysis: SFT vs. SFT + DPO (Average)}
\label{tab:dpo_robustness}
\end{table}

These results suggest that DPO enhances \textbf{generalization}, especially on unseen or more diverse test cases, even with a relatively small amount of preference training data. From a data perspective, this aligns with the core insight of LIMA~\cite{zhou2023limaalignment}---that a few high-quality preference samples can be highly effective for alignment.

In our case, although the total amount of DPO data is limited (8k preference pairs derived from 20k samples), it still results in noticeable improvements in both robustness and generalization for text-to-SQL tasks. This underscores that data quality and diversity are key to effective model tuning. We attribute the quality of our DPO data to the execution-based verification method used during preference construction.

\subsection{Impact of Data Translation and Augmentation Strategies}
\label{sec:data_translation_augmentation}

To clarify the impact of different data translation and augmentation strategies on the performance of our Supervised Fine-tuning (SFT) baselines, we provide a comparison of SFT results under three distinct approaches:

\begin{itemize}
    \item \textbf{Translation (once):} One-pass LLM translation without any further refinement (as described in Section 3.2 of the main paper).
    \item \textbf{Translation (iterative):} LLM translation enhanced with execution feedback within an iterative loop (as detailed in Section 3.2 of the main paper).
    \item \textbf{Translation + Augmented data:} Combines the translated data with newly generated question-SQL pairs derived from table rows (as described in Section 3.3 of the main paper). The \textbf{Final} setting integrates iterative refinement, data augmentation, and Direct Preference Optimization (DPO).
\end{itemize}

The performance of these strategies on the PostgreSQL and MySQL dialects is summarized in Table \ref{tab:translation_augmentation_impact}.

\begin{table}[h!]
\centering
\renewcommand{\arraystretch}{1.2}
\begin{tabular}{lcc}
\hline
Strategy & PostgreSQL & MySQL \\
\hline
Translation (once) & 63.49 & 60.09 \\
Translation (iterative w/ execution) & 69.97 & 67.63 \\
Translation + Augmented data & 71.36 & 70.34 \\
Final & 71.98 & 72.86 \\
\hline
\end{tabular}
\vskip 0.06in
\caption{Impact of Data Translation and Augmentation Strategies on Performance}
\label{tab:translation_augmentation_impact}
\end{table}

These results clearly demonstrate the significant benefits of incorporating execution feedback in the translation process and further enhancing the training data through augmentation techniques. The "Final" setting, which combines iterative refinement, data augmentation, and DPO, achieves the highest performance on both PostgreSQL and MySQL.

\subsection{Impact of Multi-Dialect Training}
\label{sec:multi_dialect_training}

To investigate how well Large Language Models (LLMs) can learn from training data across different SQL dialects and the potential benefits of multi-dialect training, we conducted a Supervised Fine-tuning (SFT) experiment using the Spider dataset with two primary dialects: PostgreSQL and MySQL. We evaluated the cross-dialect generalization performance under three different training settings:

\begin{itemize}
    \item \textbf{Only PostgreSQL:} Model trained on Spider data augmented with PostgreSQL-specific syntax.
    \item \textbf{Only MySQL:} Model trained on Spider data augmented with MySQL-specific syntax.
    \item \textbf{Mixed Training:} Model trained on a dataset combining both PostgreSQL and MySQL augmented Spider data.
\end{itemize}

The results of this experiment are summarized in Table \ref{tab:multi_dialect_results}.

\begin{table}[h!]
\centering
\renewcommand{\arraystretch}{1.2}
\begin{tabular}{lccc}
\hline
Training Setting & PostgreSQL & MySQL & Average \\
\hline
Only PostgreSQL & 74.94 & 58.30 & 66.62 \\
Only MySQL & 59.94 & 74.96 & 67.45 \\
Mixed Training & 72.15 & 68.61 & 70.38 \\
\hline
\end{tabular}
\vskip 0.06in
\caption{Cross-Dialect Generalization Performance}
\label{tab:multi_dialect_results}
\end{table}

We observed two interesting trends from these results:

\begin{itemize}
    \item Models tend to overfit to the specific syntax they are trained on, resulting in a significant performance drop when evaluated on a different dialect.
    \item Although a naive mixed training approach improves the overall average performance, it slightly reduces the peak performance achieved on individual dialects when trained solely on that dialect.
\end{itemize}

We hypothesize that this phenomenon is related to "forgetting"~\cite{Kirkpatrick_2017, alexandrov2024mitigatingcatastrophicforgettinglanguage}. To further improve cross-dialect generalization, more sophisticated mixing strategies such as in-batch mixing~\cite{pan2024scalebioscalablebileveloptimization, xie2023doremioptimizingdatamixtures}, data replay~\cite{lin2024mitigatingalignmenttaxrlhf}, or even model merging~\cite{alexandrov2024mitigatingcatastrophicforgettinglanguage} may be necessary.

\subsection{Empirical Analysis of Data Diversity}
\label{appendix:data_diversity}
We present an empirical experiment designed to investigate the data diversity of our generated samples compared to a baseline. As discussed in the main body of the paper, the validity verification mechanism based solely on SQL execution correctness during the data synthesis phase may suffer from dimensionality limitations. This can potentially lead Large Language Models (LLMs) to generate structurally homogeneous question-answer (Q\&A) pairs, consequently reducing data diversity.

To address this potential diversity collapse, our generative iteration process explicitly incorporates diversity by varying the prompts with in-context exemplars at every generation round, a strategy similar to self-instruct~\cite{wang-etal-2023-self-instruct} 

To empirically study the data diversity issue, we conducted the following experiment:
\begin{table}[ht]
\centering
\renewcommand{\arraystretch}{1.2}
\begin{tabular}{lc}
\hline
Dataset & Similarity Score \\
\hline
Our Data & 0.470 \\
Spider Sample Comparison & 0.672 \\
\hline
\end{tabular}
\vskip 0.06in
\caption{Comparison of Cosine Similarity Scores (TF-IDF Embeddings) between Our Generated Data and Baseline Spider Samples. Our data (10K samples paired with 5K Spider samples) shows a lower average similarity score compared to the similarity within the Spider dataset (10K Spider samples paired with highest similarity HumanEval samples), suggesting higher diversity.}
\label{tab:diversity_comparison}
\end{table}

The results presented in Table \ref{tab:diversity_comparison} show that our generated data is more diverse than the baseline Spider samples, suggesting that our multi-round varying prompting strategy effectively mitigates diversity collapse.

On the other hand, we believe the most effective way to ensure data diversity is through access to diverse database schemas. Our method focuses on generating accurate and precise QA pairs given a particular database. As more varied databases become available, our generation framework is expected to produce even more diverse and useful training data.

\subsection{Generalization to Single-Domain Datasets}
\label{sec:single_domain_generalization}

Our method is not limited by dataset type and can be applied to single-domain datasets as well. To demonstrate this, we applied our model to the MimicSQL dataset~\cite{wang2020texttosqlgenerationquestionanswering, deng-etal-2022-recent} using the MySQL dialect \textbf{without further training}. The results are shown in Table \ref{tab:mimicsql_accuracy}.

\begin{table}[h!]
\centering
\renewcommand{\arraystretch}{1.2}
\begin{tabular}{lc}
\hline
Model & Accuracy \\
\hline
GPT-4o & 72.87 \\
DeepSeek-Coder-7B & 63.66 \\
Qwen2.5-Coder-7B & 61.46 \\
StructLM-7B & 38.34 \\
ExeSQL (ours) & \textbf{76.07} \\
\hline
\end{tabular}
\vskip 0.06in
\caption{Accuracy on the MimicSQL Dataset (MySQL Dialect, Zero-Shot)}
\label{tab:mimicsql_accuracy}
\end{table}
These results show that our method generalizes well to the single-domain setting, achieving a competitive accuracy of 76.07\% on the MimicSQL dataset in a zero-shot manner, even outperforming larger, general-purpose models like GPT-4o. This highlights the robustness and adaptability of our approach beyond cross-domain benchmarks.

\subsection{Details of Evaluation Datasets}
\label{appendix:eval_data}

\begin{table}[ht]
    \centering
    \renewcommand{\arraystretch}{1.2}
    \setlength{\tabcolsep}{5pt}
    
    \label{tab:dataset_samples}
    \begin{tabular}{lccccc}
        \toprule
        \textbf{Dataset} & \textbf{Spider} & \textbf{WikiSQL} & \textbf{Dr.Spider} & \textbf{Bird} & \textbf{MimicSQL}\\
        \midrule
        \# samples & 2,147 & 8,421 & 15,269 & 1534 & 999\\
        \bottomrule
    \end{tabular}
    \vskip 0.06in
    \caption{Number of samples in different datasets}
\end{table}

\textbf{Spider.} Spider provides a diverse collection of training and development samples, along with a hidden test set. The training set includes a mix of manually annotated examples and additional samples sourced from previous text-to-SQL datasets. Covering a wide range of databases across various domains, Spider serves as a comprehensive benchmark for evaluating cross-domain text-to-SQL performance. We used the test set of Spider with 2,147 examples to perform evaluation here.

\textbf{WikiSQL.} WikiSQL is a large-scale dataset consisting of natural language questions, SQL queries, and structured tables extracted from Wikipedia. It offers a well-organized set of training, development, and test examples, each containing a question, a table, an SQL query, and the expected execution result. We used the dev set of Spider with 8,421 examples to perform evaluation here.

\textbf{Dr.Spider.} Dr.Spider, an extension of Spider, introduces various perturbations across questions, databases, and SQL queries to assess the robustness of text-to-SQL models. It includes test sets designed to evaluate the impact of database modifications, question variations, and SQL transformations, making it a challenging benchmark for robustness testing. We used the perturbed set over all the questions, databases, and SQL queries with 15,269 examples to perform evaluation here. Detailed perturbation types are shown in Table~\ref{tab:perturb_samples}.

\begin{table}[ht]
    \centering
    \renewcommand{\arraystretch}{1.2}
    \setlength{\tabcolsep}{8pt}
    \begin{tabular}{lc}
        \toprule
        \textbf{Perturb Type} & \textbf{\# Samples} \\
        \midrule
        DB\_DBcontent\_equivalence & 382 \\
        DB\_schema\_abbreviation & 2853 \\
        DB\_schema\_synonym & 2619 \\
        NLQ\_column\_attribute & 119 \\
        NLQ\_column\_carrier & 579 \\
        NLQ\_column\_synonym & 563 \\
        NLQ\_column\_value & 304 \\
        NLQ\_keyword\_carrier & 399 \\
        NLQ\_keyword\_synonym & 953 \\
        NLQ\_multitype & 1351 \\
        NLQ\_others & 2819 \\
        NLQ\_value\_synonym & 506 \\
        SQL\_comparison & 178 \\
        SQL\_DB\_number & 410 \\
        SQL\_DB\_text & 911 \\
        SQL\_NonDB\_number & 131 \\
        SQL\_sort\_order & 192 \\
        \bottomrule
    \end{tabular}
    \vskip 0.06in
    \caption{Perturbation Types and Sample Counts}
    \label{tab:perturb_samples}
\end{table}

\textbf{Bird.} Bird is a large-scale, challenging dataset specifically designed to evaluate the in-context learning capabilities of text-to-SQL models. It encompasses a wide variety of complex SQL queries and database schemas, demanding strong reasoning and schema understanding. The dataset includes training, development, and test splits. For our experiments, we utilized the development set of BIRD, which comprises 1,534 examples for evaluation.

\textbf{MimicSQL.} MimicSQL is a single-domain text-to-SQL dataset derived from the MIMIC-III electronic health records database. It focuses on the medical domain, presenting unique challenges related to medical terminology and complex database structures within healthcare. The dataset includes training and test sets. We performed our evaluation on the test set of MimicSQL\_natural, which contains 999 examples.

\subsection{Limitations of Rule-Based SQL Transpilers}
\label{appendix:Rule_base_limitations}
Static analysis and syntax-based SQL transpiler (Rule-based transpiler) is an interesting direction for dialect SQL generation tasks. However, our observations highlight several limitations that make this approach less desirable compared to methods leveraging execution feedback.
\subsubsection*{Observation 1: Rule-Based Transpilers Still Require Execution Feedback}
While tools like SQLGlot provide syntax-level SQL transpilation, they cannot guarantee semantic correctness or executable validity in the target dialect. As shown in Table \ref{tab:sqlglot_error}, in many cases, SQLGlot generates syntactically valid but semantically incorrect queries, making execution feedback still necessary for validation and refinement.
\begin{table}[ht]
\centering
\begin{tabular}{p{3.5cm} p{12cm}}
\hline
\textbf{Aspect} & \textbf{Content} \\
\hline
SQLGlot Output & \texttt{SELECT T1.rating\_score, T2.director\_name FROM ratings AS T1 JOIN movies AS T2 ON T1.movie\_id = T2.movie\_id WHERE T2.movie\_title = 'When Will I Be Loved'} \\
\hline
Execution Error & Error 1140 (42000): In aggregated query without GROUP BY, expression \#2 of SELECT list contains nonaggregated column 'movie\_platform.T2.director\_name'; this is incompatible with sql\_mode=only\_full\_group\_by \\
\hline
Original SQLite Query & \texttt{SELECT T1.rating\_score, T2.director\_name FROM ratings AS T1 JOIN movies AS T2 ON T1.movie\_id = T2.movie\_id WHERE T2.movie\_title = 'When Will I Be Loved'} \\
\hline
Correct MySQL Query & \texttt{SELECT avg(T1.rating\_score) AS average\_rating, T2.director\_name FROM ratings AS T1 JOIN movies AS T2 ON T1.movie\_id = T2.movie\_id WHERE T2.movie\_title = 'When Will I Be Loved' GROUP BY T2.director\_name} \\
\hline
\end{tabular}
\vskip 0.06in
\caption{SQLGlot misses the required \texttt{GROUP BY} clause, which causes execution failure in MySQL under strict SQL modes.}
\label{tab:sqlglot_error}
\end{table}
\subsubsection*{Observation 2: Rule-Based Methods Can Hardly Do Multi-Round Refinement}
Our method supports iterative refinement by injecting failing case inputs to the next round's prompt (similar to self-correction). However, rule-based transpilers like SQLGlot require manual updates from programming experts to improve over iterations, making them less adaptable in practice.
\subsubsection*{Observation 3: Rule-Based Performance Is Worse}
We compared the performance of SQLGlot against a single-round LLM-based translation.
\begin{table}[ht]
\centering
\renewcommand{\arraystretch}{1.2}
\begin{tabular}{lc}
\hline
Method & Accuracy (\%) \\
\hline
LLM (API, 1 round) & 56.32 \\
SQLGlot & 35.18 \\
\hline
\end{tabular}
\vskip 0.06in
\caption{Accuracy Comparison: Rule-Based vs. LLM-Based Translation}
\label{tab:rule_based_vs_llm}
\end{table}

Table \ref{tab:rule_based_vs_llm} shows a clear gap in accuracy, further highlighting the limitations of relying solely on static transpilation.
\subsubsection*{Observation 4: Combine Usage of Rule-Based Transpiler and LLM Can Reduce LLM Call Cost}
We also analyzed whether pre-filtering with SQLGlot can reduce the total number of LLM calls. Assuming SQLGlot correctly solves 35\% of queries, and the LLM solves 56\% of the remaining ones in each round (up to 3 rounds), the estimated number of API calls is shown in Table \ref{tab:llm_call_cost}. Pre-filtering with SQLGlot results in roughly 35\% savings in LLM calls, consistent with the success rate of SQLGlot. However, while this combination can reduce costs, it still necessitates the use of an LLM and does not overcome the fundamental limitations in semantic correctness and iterative refinement of purely rule-based approaches.
\begin{table}[ht]
\centering
\setlength{\tabcolsep}{4pt}
\renewcommand{\arraystretch}{1.2}
\begin{tabular}{lccc}
\hline
Setting & \multicolumn{3}{c}{LLM Calls} \\
\cline{2-4}
 & 1k & 10k & 100k \\
\hline
LLM-only & 1,634 & 16,336 & 163,360 \\
SQLGlot + LLM & 1,062 & 10,618 & 106,184 \\
\hline
\end{tabular}
\vskip 0.06in
\caption{Estimated LLM API Calls with and without SQLGlot Pre-filtering for 3 rounds}
\label{tab:llm_call_cost}
\end{table}
\subsection{Execution feedback efficiency}
\label{appendix:Execution_feedback_efficiency}
To better understand the potential efficiency bottlenecks, we analyze execution time across databases of varying complexity using the MySQL engine. Specifically, we compare execution performance on the Spider dataset (with relatively small and simple databases) and the BIRD dataset \cite{li2023can} (which contains significantly larger and more complex schemas). The complexity of database can be shown in Table \ref{tab:Execution Time Analysis} through the “Avg. Rows per DB” metric.

We acknowledge that execution-based validation introduces overhead, but we argue that the cost remains acceptable, especially given the significant gains in data quality and model generalization. Furthermore, since execution feedback is only used during data generation (not inference), this cost is one-time and offline. Future improvements could involve caching, schema-aware pruning, or batched execution to further enhance scalability.
\begin{table*}[t]
    \centering
    \renewcommand{\arraystretch}{1.2}
    \setlength{\tabcolsep}{4pt}
    \resizebox{\textwidth}{!}{
    \begin{tabular}{lcccc}
        \hline
        Dataset & Avg. Rows per DB & Avg. Import Time & Avg. Execution Time & Time for processing 8,000 Samples \\
        \hline
        Spider (small-scale) & 5,910.3 & 0.233 s & 0.00618 s & 87.2 s \\
        BIRD (large-scale scenarios) & 256,231 & 21.647 s & 0.06149 s & 1,944 s \\
        Training (7B model) & - & - & - & **11,100s** \\
        \hline
    \end{tabular}
    }
    \caption{Execution Time Analysis (MySQL)}
    \label{tab:Execution Time Analysis}
\end{table*}

\subsection{Answer Extraction}
\label{appendix:answer_extraction}
Since LLM-based text-to-SQL generation often introduces variance, such as generating unrelated information or placing answers in undesired formats, we incorporate a regex-based answer extraction tool for robust evaluation. Common formatting issues include repeated questions, answers enclosed in code blocks (e.g., \texttt{``...''}), and additional explanations.

\subsection{Detailed Translation-based Bootstrapping Process}
\label{appendix:translation}
The original Spider dataset is based on SQLite SQL. We used GPT-4o API to generate MySQL and PostgreSQL SQL queries based on the given SQLite SQL, natural language questions, and table information (including table names and column names). The generation process followed a structured approach to ensure high accuracy and compatibility across SQL dialects.
In \ref{tab:prompt_example}, \ref{tab:prompt_example_mysql}, we describe the prompt used for GPT-4o, which highlights key differences between SQLite SQL and PostgreSQL/Mysql SQL. The prompt also provides several input-output examples that illustrate how SQLite SQL should be transformed into the target SQL dialects. These examples help GPT-4o understand the conversion rules and adapt the syntax accordingly.

\begin{figure*}[ht] % 使用 figure* 环境允许图片跨双栏
    \centering % 居中整个 figure 环境
    \begin{minipage}[t]{0.48\textwidth} % 第一个图片的小页面
        \centering
        \includegraphics[width=\linewidth]{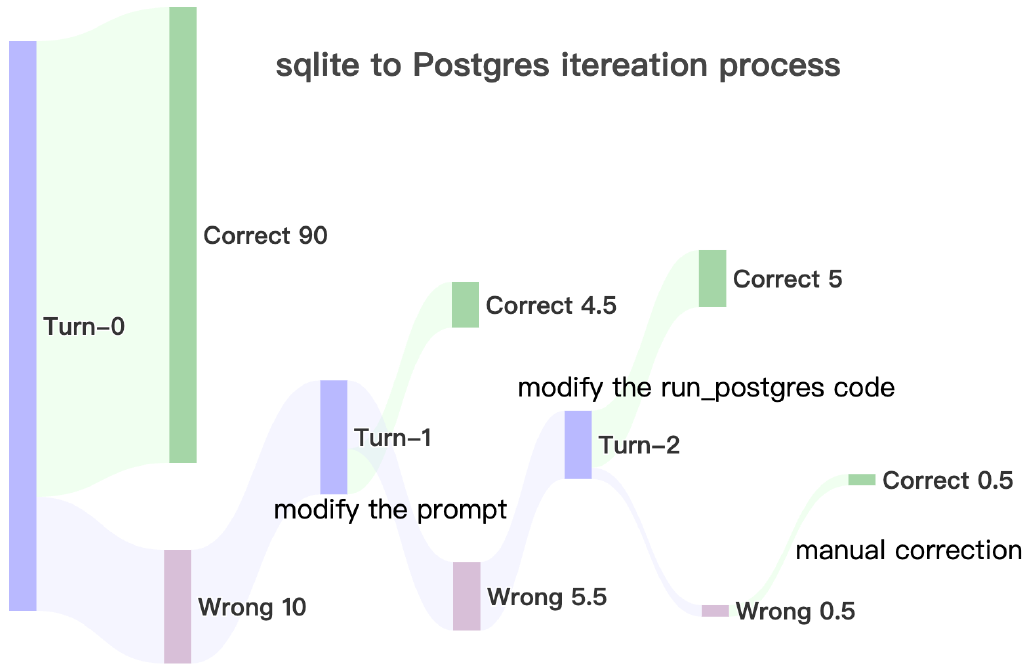}
        % 如果需要子标题，可以使用 \subcaption{...}，但这里要求不单独加caption
        \caption{SQLite to PostgreSQL process} % 可以用这种方式给图片加无编号的标题
        \label{fig:postgres_iteration} % 单独的label仍然可以保留
    \end{minipage}
    \hfill % 在两个 minipage 之间添加水平空间
    \begin{minipage}[t]{0.48\textwidth} % 第二个图片的小页面
        \centering
        \includegraphics[width=\linewidth]{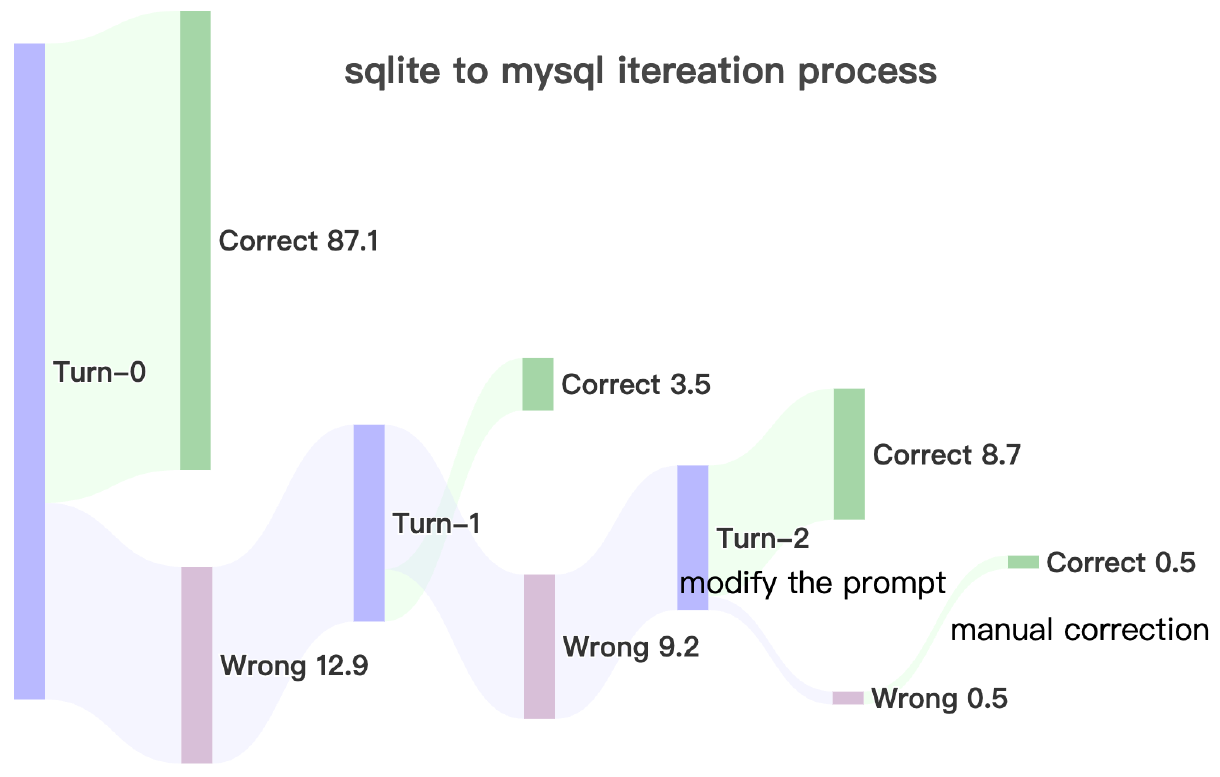}
        \caption{SQLite to MySQL process} % 可以用这种方式给图片加无编号的标题
        \label{fig:mysql_iteration} % 单独的label仍然可以保留
    \end{minipage}
    % \caption{Process of SQL translation and iteration for PostgreSQL and MySQL dialects.} % 整个组合的总标题
    \label{fig:translation_processes} % 整个组合的总label
\end{figure*}

\paragraph{PostgreSQL Generation Process}
Figure \ref{fig:postgres_iteration} illustrates the process of generating PostgreSQL SQL.
In the first iteration of PostgreSQL generation, we found that around 680 queries failed with compilation errors. To address this, we enhanced the prompt by including additional PostgreSQL-specific features and updated the input-output examples with corrected versions of some failed queries from the first iteration (e.g., Ensure all `JOIN` operations explicitly specify the `ON` condition; When using GROUP BY, all selected non-aggregated columns must be explicitly listed in the GROUP BY clause). After applying the modified prompt, the number of incorrect queries decreased to about 400. Upon reviewing the errors, we discovered that most issues were caused by tables containing cells with improper formats (e.g., empty cells or invalid values like ""). To resolve this, we adjusted the code responsible for running PostgreSQL queries by skipping rows with problematic data during the conversion process. After implementing the data-cleaning step, only 30 queries remained incorrect. These were corrected manually to achieve a fully accurate PostgreSQL SQL dataset.

% \begin{figure}[ht]
%     % \centering
%     \includegraphics[width=0.5\textwidth]{figures/sql2mysql.pdf}
%     \caption{sqlite to mysql process}
%     \label{fig:mysql_iteration}
% \end{figure}

\paragraph{Mysql Generation Process}
As shown in Figure \ref{fig:mysql_iteration}, the MySQL generation process followed a similar iterative approach.
In the first iteration, GPT-4o generated MySQL SQL for all queries. After evaluation, we found that approximately 890 queries were incorrect due to compilation errors. Since the number of errors is still very high, we repeat the same iteration based on the incorrect samples. After the second iteration, there are still 630 examples that remain incorrect.  To improve accuracy, we refined the prompt, adding more MySQL-specific features and incorporating corrected versions of some failed queries from the first two iterations. We asked GPT-4o to use backquotes for table names or column names when necessary. With the improved prompt, the number of incorrect queries dropped to around 20. Finally, we manually corrected these remaining 20 queries to achieve a fully accurate MySQL SQL dataset.  

\subsection{Handling Complex SQL Features}
\label{sec:complex_sql_features}
Unlike static analysis or rule-based transpilers, our method leverages both SQL and question semantics. This allows it to better preserve and adapt complex functional behavior during translation. Below, Table \ref{tab:translation_over} and Table \ref{tab:translation_ilike} present examples demonstrating the capability of our method to correctly handle PostgreSQL-specific features, addressing the limitations often encountered by simpler translation approaches.
\begin{table*}[t]
\centering
\begin{tabular}{p{0.11\linewidth} p{0.4\linewidth} p{0.4\linewidth}}
\hline
\textbf{Example} & \textbf{SQLite Query} & \textbf{Translated PostgreSQL} \\
\hline
Example 1 (No OVER in SQLite) &
\begin{lstlisting}[language=SQL, basicstyle=\ttfamily\footnotesize, breaklines=true]
SELECT T2.director_name FROM ratings AS T1 INNER JOIN movies AS T2 ON T1.movie_id = T2.movie_id
WHERE T2.release_year BETWEEN 1960 AND 1985 GROUP BY T2.director_name HAVING COUNT(T2.movie_id) > 10
ORDER BY T2.director_name LIMIT 2
\end{lstlisting}
&
\begin{lstlisting}[language=SQL, basicstyle=\ttfamily\footnotesize, breaklines=true]
SELECT director_name, movie_title FROM (SELECT m.director_name,
m.movie_title, m.movie_id, RANK() OVER (PARTITION BY m.director_id ORDER BY
COUNT(r.rating_id) FILTER (WHERE r.rating_score::FLOAT > 5) DESC) as rank
FROM movies AS m JOIN ratings r ON m.movie_id = r.movie_id WHERE
m.release_year BETWEEN 1960 AND 1985 GROUP BY m.director_id,
m.movie_id, m.director_name, m.movie_title) subquery WHERE rank::INTEGER = 1
AND director_name IN (SELECT director_name FROM movies WHERE
release_year::INTEGER BETWEEN 1960 AND 1985 GROUP BY director_name
HAVING COUNT(movie_id) >= 10) ORDER BY director_name LIMIT 2
\end{lstlisting}
\\
\hline
Example 2 (Window used) &
\begin{lstlisting}[language=SQL, basicstyle=\ttfamily\footnotesize, breaklines=true]
SELECT T3.years, T3.episode_id FROM (SELECT DISTINCT T2.years AS years, T2.episode_id, row_number() OVER (PARTITION BY
(Window T2.episode_id ORDER BY T2.year) AS rn FROM person AS T1 INNER JOIN Award AS T2 ON T1.person_id = T2.person_id WHERE T2.award =
used) 'Television' AND T2.award_category = 'Silver Gavel Award' AND T1.name = 'Constantine Makris' AND T2.result = 'Winner' AND
T2.organization = 'American Bar Association Silver Gavel Awards for Media and the Arts') AS T3 GROUP BY T3.episode_id HAVING
COUNT(T3.years - T3.rn) >= 2
\end{lstlisting}
&
\begin{lstlisting}[language=SQL, basicstyle=\ttfamily\footnotesize, breaklines=true]
SELECT T3.years, T3.episode_id FROM (SELECT DISTINCT T2.years AS
years, T2.episode_id, row_number() OVER (PARTITION BY T2.episode_id ORDER BY
T2.year) AS rn FROM person AS T1 JOIN Award AS T2 ON T1.person_id =
T2.person_id WHERE T2.award = 'Television' AND T2.award_category = 'Silver
Gavel Award' AND T1.name = 'Constantine Makris' AND T2.result = 'Winner' AND
T2.organization = 'American Bar Association Silver Gavel Awards for Media and
the Arts') AS T3 GROUP BY T3.episode_id, T3.years HAVING COUNT(T3.years -
T3.rn::FLOAT) >= 2
\end{lstlisting}
\\
\hline
\end{tabular}
\caption{Translation Examples: PostgreSQL \texttt{OVER} (Window Function)}
\label{tab:translation_over}
\end{table*}

\begin{table*}[t]
\centering
\begin{tabular}{p{0.115\linewidth} p{0.4\linewidth} p{0.4\linewidth}}
\hline
\textbf{Example} & \textbf{SQLite Query} & \textbf{Translated PostgreSQL} \\
\hline
Example 3 (LIKE on date) &
\begin{lstlisting}[language=SQL, basicstyle=\ttfamily\footnotesize, breaklines=true]
SELECT T3.keyword_name FROM movie AS T1 INNER JOIN movie_keyword AS T2 ON T1.movie_id = T2.movie_id INNER JOIN keyword
AS T3 ON T2.keyword_id = T3.keyword_id WHERE T1.release_date BETWEEN '2006-01-01' AND '2006-12-31' GROUP BY T3.keyword_name
ORDER BY COUNT(T3.keyword_name) DESC LIMIT 1
\end{lstlisting}
&
\begin{lstlisting}[language=SQL, basicstyle=\ttfamily\footnotesize, breaklines=true]
SELECT T3.keyword_name FROM movie AS T1 JOIN movie_keyword AS T2 ON T1.movie_id = T2.movie_id JOIN keyword
AS T3 ON T2.keyword_id = T3.keyword_id WHERE T1.release_date BETWEEN '2006-01-01' AND '2006-12-31' GROUP BY T3.keyword_name
ORDER BY COUNT(T3.keyword_name) DESC LIMIT 1
\end{lstlisting}
\\
\hline
Example 4 (Regex-like keyword match) &
\begin{lstlisting}[language=SQL, basicstyle=\ttfamily\footnotesize, breaklines=true]
SELECT keyword_name FROM keyword WHERE keyword_name LIKE '%christmas%'
\end{lstlisting}
&
\begin{lstlisting}[language=SQL, basicstyle=\ttfamily\footnotesize, breaklines=true]
SELECT keyword_name FROM keyword WHERE keyword_name ILIKE '%christmas%'
\end{lstlisting}
\\
\hline
\end{tabular}
\caption{Translation Examples: PostgreSQL \texttt{ILIKE} (RegEx Like)}
\label{tab:translation_ilike}
\end{table*}

\subsection{Example of Execution-based Rejection Sampling}
\label{appendix:case_study_rejection_sampling}

Table \ref{tab:sampling_example} presents an example of the proposed execution-based rejection sampling, where the question originates from the WikiSQL training set. We reformat the question to match the format used for our training and applied best-of-8 sampling with a cold-started model. In the upper section, we illustrate an example of generating new training data by sampling the correct Postgres SQL query. Among the eight generated queries, only one correctly queries the database using "date='January 16'", aligning with the question's description. Although the other queries are semantically similar and syntactically correct, they fail to retrieve the correct answer.

The lower section provides an example of sampling for DPO training data. The question asks about 'scored 24 points,' whereas the rejected answer retrieves values where the score is greater than 24, which is clearly incorrect.

\begin{table*}[t]\centering
\begin{minipage}{1.0\textwidth}\vspace{0mm}    \centering
\begin{sectionbox}[]{Data Instance}
    \centering
      \footnotesize
    \begin{tabular}{p{0.97\textwidth} c}
\textbf{Best of N}\\
\vspace{1pt}
[Input]: \\
You need to generate a Postgres SQL based on the following question and table information. Question: Who is the Opponent on January 16?  Table and columns information: table\_2\_16946097\_6: Date, H\_A\_N, Opponent, Score, Record. \\

\vspace{2pt}

[Outputs]: \\
SELECT Opponent FROM table\_2\_16946097\_6 WHERE Date = 'January 16' \\
SELECT Opponent FROM table\_2\_16946097\_6 WHERE Date = 'Jan 16' \\
SELECT Opponent FROM table\_2\_16946097\_6 WHERE Date = 'Jan 16' \\
SELECT Opponent FROM table\_2\_16946097\_6 WHERE Date = 'Jan 16' \\
SELECT Opponent FROM table\_2\_16946097\_6 WHERE Date = 'Jan 16' \\
SELECT Opponent FROM table\_2\_16946097\_6 WHERE Date = '1/16' \\
SELECT Opponent FROM table\_2\_16946097\_6 WHERE Date = 'Jan 16' \\
SELECT Opponent FROM table\_2\_16946097\_6 WHERE Date = '1/16' \\

\vspace{2pt}

[Correct Answer]: \\
SELECT Opponent FROM table\_2\_16946097\_6 WHERE Date = 'January 16' \\
\vspace{10pt}

\textbf{Preference Pair}\\
\vspace{1pt}
[Input]: \\
You need to generate a Postgres SQL based on the following question and table information. Question: What was the record after the game in which the Hurricanes scored 24 points?  Table and columns information: table\_1\_20928682\_1: Game, Date, Opponent, Result, Hurricanes\_points, Opponents, Record. \\
\vspace{2pt}
[Chosen Answer]: \\
SELECT table\_1\_20928682\_1.Record FROM table\_1\_20928682\_1\\
\hspace{8mm} WHERE table\_1\_20928682\_1.Hurricanes\_points::FLOAT = 24 \\
\vspace{2pt}
[Rejected Answer]: \\
SELECT table\_1\_20928682\_1.Record FROM table\_1\_20928682\_1\\
\hspace{8mm} WHERE table\_1\_20928682\_1.Hurricanes\_points::FLOAT > 24 \\
\vspace{1pt}

    \end{tabular}
\end{sectionbox}
\vspace{-2mm}
\caption{Data instance of our iteration.}
    \label{tab:sampling_example}
\end{minipage}
\end{table*}

\begin{table*}[t]\centering
\begin{minipage}{1.0\textwidth}\vspace{0mm}    \centering
\begin{sectionbox}[]{Prompt Format for SQLite to PostgreSQL Conversion}
    \centering
      \footnotesize
    \begin{tabular}{p{0.97\textwidth}}
\textbf{Prompt Description:} \\
You are an expert in SQL conversion. Convert SQLite SQL statements to PostgreSQL SQL while strictly following PostgreSQL's syntax. \\

\textbf{Important Instructions:} \\
1. \textbf{Input Format:} Each line in the input file follows this format: \\
\hspace{4mm} SQLite SQL \textbackslash t db\_id \\
\hspace{4mm} Example: \texttt{SELECT count(*) FROM head WHERE age > 56 department\_management} \\
\hspace{8mm} - \texttt{"SELECT count(*) FROM head WHERE age > 56"} is the SQLite SQL. \\
\hspace{8mm} - \texttt{"department\_management"} is the db\_id. \\

2. \textbf{Output Format (STRICTLY ENFORCED):} \\
\hspace{4mm} You must return the converted PostgreSQL SQL followed by the same db\_id as input: \\
\hspace{4mm} Example: \texttt{SELECT count(*) FROM head WHERE head.age::INTEGER > 56 department\_management} \\

\textbf{Rules to Follow:} \\
- Do not add explanations, comments, or any extra text. \\
- Output must be one line per input, separated by a single `\textbackslash t`. \\
- For column names, add the table name before each column in PostgreSQL SQL. \\
- Ensure db\_id remains exactly as in the input. \\
- Ensure explicit column references (e.g., \texttt{table.column}). \\
- When using GROUP BY, all selected non-aggregated columns must be explicitly listed in the GROUP BY clause to avoid errors in PostgreSQL. \\
- Ensure all `JOIN` operations explicitly specify the `ON` condition. Avoid using implicit joins or missing `ON` conditions, as PostgreSQL requires explicitly defined relationships between tables.\\
- If a table has an alias in the `FROM` or `JOIN` clause, always use the alias instead of the original table name in `SELECT`, `WHERE`, and other clauses.\\
- for SELECT DISTINCT, ORDER BY expressions must appear in select list\\
- Ensure that tables are referenced in `JOIN` statements in the correct order: a table must be defined before being used in an `ON` condition.\\

\vspace{5pt}
\textbf{Example 1:} \\
\textbf{Input:} \\
\hspace{4mm} \texttt{SELECT DISTINCT T1.player\_name, T1.birthday FROM Player AS T1 JOIN Player\_Attributes AS T2 ON T1.player\_api\_id = T2.player\_api\_id ORDER BY potential DESC LIMIT 5} \texttt{soccer\_1} \\

\textbf{Output:} \\
\hspace{4mm} \texttt{SELECT DISTINCT T1.player\_name, T1.birthday, T2.potential::FLOAT FROM Player AS T1 JOIN Player\_Attributes AS T2 ON T1.player\_api\_id = T2.player\_api\_id ORDER BY T2.potential::FLOAT DESC LIMIT 5} \texttt{soccer\_1} \\

\vspace{2pt}
\textbf{Example 2:} \\
\textbf{Input:} \\
\hspace{4mm} SELECT count(*) FROM head WHERE age > 56 \texttt{department\_management} \\
\textbf{Output:} \\
\hspace{4mm} SELECT count(*) FROM head WHERE head.age::INTEGER > 56 \texttt{department\_management} \\

\vspace{2pt}
\textbf{Example 3:} \\
\textbf{Input:} \\
\hspace{4mm} SELECT avg(lat), avg(long) FROM station WHERE city = "San Jose" \texttt{bike\_1} \\
\textbf{Output:} \\
\hspace{4mm} SELECT AVG(lat::FLOAT), AVG(long::FLOAT) FROM station WHERE station.city = "San Jose" \texttt{bike\_1} \\

\vspace{2pt}
\textbf{Example 4:} \\
\textbf{Input:} \\
\hspace{4mm} SELECT T1.age FROM Person AS T1 JOIN PersonFriend AS T2 ON T1.name  =  T2.friend WHERE T2.name =  'Zach' AND T2.year  =  (SELECT max(YEAR) FROM PersonFriend WHERE name =  'Zach') \texttt{network\_2} \\
\textbf{Output:} \\
\hspace{4mm} SELECT T1.age FROM Person AS T1 JOIN PersonFriend AS T2 ON T1.name  =  T2.friend WHERE T2.name =  'Zach' AND T2.year::FLOAT =  (SELECT MAX(YEAR::FLOAT) FROM PersonFriend WHERE name =  'Zach') \texttt{network\_2} \\

\vspace{5pt}
\textbf{Now, convert the following SQLite SQL to PostgreSQL SQL. Output strictly in format: SQL \textbackslash t db\_id.} \\

    \end{tabular}
\end{sectionbox}
\vspace{-2mm}
\caption{Prompt example for converting SQLite SQL to PostgreSQL SQL.}
    \label{tab:prompt_example}
\end{minipage}
\end{table*}

\begin{table*}[t]\centering
\begin{minipage}{1.0\textwidth}\vspace{0mm}    \centering
\begin{sectionbox}[]{Prompt Format for SQLite to MySQL Conversion}
    \centering
      \footnotesize
    \begin{tabular}{p{0.97\textwidth}}
\textbf{Prompt Description:} \\
You are an expert in SQL conversion. Convert SQLite SQL statements to MySQL SQL while strictly following MySQL's syntax. \\

\textbf{Important Instructions:} \\
1. \textbf{Input Format:} Each line in the input file follows this format: \\
\hspace{4mm} Index \textbackslash t SQLite SQL \textbackslash t db\_id \\
\hspace{4mm} Example: \texttt{1 SELECT T3.course\_name , count(*) FROM students AS T1 JOIN student\_course\_registrations AS T2 ON T1.student\_id = T2.student\_id JOIN courses AS T3 ON T2.course\_id = T3.course\_id GROUP BY T2.course\_id \ student\_assessment} \\

2. \textbf{Output Format (STRICTLY ENFORCED):} \\
\hspace{4mm} Just output the converted MySQL SQL query (do not include index or db\_id). \\
\hspace{4mm} Example: \texttt{SELECT T3.course\_name , count(*) FROM Students AS T1 JOIN Student\_Course\_Registrations AS T2 ON T1.student\_id = T2.student\_id JOIN Courses AS T3 ON T2.course\_id = T3.course\_id GROUP BY T2.course\_id, T3.course\_name} \\

\textbf{Rules to Follow:} \\
- Do not add explanations, comments, or any extra text. \\
- Output must be one line per input, separated by a single `\textbackslash t`. \\
- When using \texttt{GROUP BY}, all selected non-aggregated columns or tables must be explicitly listed in the \texttt{GROUP BY} clause to avoid errors in MySQL. \\
- When writing table names in MySQL, case matters. Refer to the provided table information to ensure correct casing. \\
- Use backquotes for table name or column name when necessary. \\
- This version of MySQL doesn't yet support 'LIMIT \& IN/ALL/ANY/SOME subquery' \\

\vspace{5pt}
\textbf{Example 1:} \\
\textbf{Input:} \\
\hspace{4mm} 2 \texttt{SELECT T1.campus , sum(T2.degrees) FROM campuses AS T1 JOIN degrees AS T2 ON T1.id = T2.campus WHERE T2.year >= 1998 AND T2.year <= 2002 GROUP BY T1.campus \ csu\_1} \\

\textbf{Output:} \\
\hspace{4mm} \texttt{SELECT T1.campus, SUM(T2.degrees) FROM Campuses AS T1 JOIN degrees AS T2 ON T1.id = T2.campus WHERE T2.year >= 1998 AND T2.year <= 2002 GROUP BY T1.campus} \\

\vspace{5pt}
\textbf{Example 2:} \\
\textbf{Input:} \\
\hspace{4mm} 3 \texttt{SELECT T1.faculty, avg(T2.salary) FROM faculties AS T1 JOIN salaries AS T2 ON T1.faculty\_id = T2.faculty\_id GROUP BY T1.faculty \ university\_pay} \\

\textbf{Output:} \\
\hspace{4mm} \texttt{SELECT T1.faculty, AVG(T2.salary) FROM Faculties AS T1 JOIN Salaries AS T2 ON T1.faculty\_id = T2.faculty\_id GROUP BY T1.faculty} \\

\vspace{5pt}
\textbf{Now, convert the following SQLite SQL to MySQL SQL. Output strictly in the format described above.} \\

    \end{tabular}
\end{sectionbox}
\vspace{-2mm}
\caption{Prompt example for converting SQLite SQL to MySQL SQL.}
    \label{tab:prompt_example_mysql}
\end{minipage}
\end{table*}

\end{document}

%% file: template.bbl
\providecommand{\CNFX}[1]{{\em{\textrm{(#1)}}}}